\documentclass[10pt,twocolumn,letterpaper]{article}
\pdfoutput=1
\usepackage[accsupp]{axessibility}
\usepackage{iccv}
\usepackage{times}
\usepackage{epsfig}
\usepackage{graphicx}
\usepackage{amsmath}
\usepackage{amssymb}
\usepackage{booktabs}
\usepackage{multirow}
\usepackage{makecell}
\usepackage{bm}
\usepackage{colortbl}
\usepackage[table]{xcolor}
\usepackage[pagebackref=true,breaklinks=true,letterpaper=true,colorlinks,bookmarks=false]{hyperref}
\usepackage[T1]{fontenc}
\usepackage{makecell}

\usepackage[capitalize]{cleveref}
\crefname{section}{Sec.}{Secs.}
\Crefname{section}{Section}{Sections}
\Crefname{table}{Table}{Tables}
\crefname{table}{Tab.}{Tabs.}

\usepackage{subfigure}
\usepackage{algorithmic}
\usepackage[ruled,lined]{algorithm2e}

\iccvfinalcopy 

\def\iccvPaperID{6103} 

\ificcvfinal\fi

\begin{document}

\title{DR-Tune: Improving Fine-tuning of Pretrained Visual Models \\by Distribution Regularization with Semantic Calibration}

\author{Nan Zhou$^{1,2}$\quad Jiaxin Chen$^{2}$\quad Di Huang$^{1,2,3}$\thanks{Corresponding author.}\\
$^{1}$State Key Laboratory of Software Development Environment, Beihang University, Beijing, China \\
$^{2}$School of Computer Science and Engineering, Beihang University, Beijing, China \\
$^{3}$Hangzhou Innovation Institute, Beihang University, Hangzhou, China \\
{\tt\small \{zhounan0431,jiaxinchen,dhuang\}@buaa.edu.cn}
}

\maketitle
\ificcvfinal\thispagestyle{empty}\fi

\begin{abstract}
   The visual models pretrained on large-scale benchmarks encode general knowledge and prove effective in building more powerful representations for downstream tasks. Most existing approaches follow the fine-tuning paradigm, either by initializing or regularizing the downstream model based on the pretrained one. The former fails to retain the knowledge in the successive fine-tuning phase, thereby prone to be over-fitting, and the latter imposes strong constraints to the weights or feature maps of the downstream model without considering semantic drift, often incurring insufficient optimization. To deal with these issues, we propose a novel fine-tuning framework, namely distribution regularization with semantic calibration (\textbf{DR-Tune}). It employs distribution regularization by enforcing the downstream task head to decrease its classification error on the pretrained feature distribution, which prevents it from over-fitting while enabling sufficient training of downstream encoders. Furthermore, to alleviate the interference by semantic drift, we develop the semantic calibration (SC) module to align the global shape and class centers of the pretrained and downstream feature distributions. Extensive experiments on widely used image classification datasets show that DR-Tune consistently improves the performance when combing with various backbones under different pretraining strategies. Code is available at: \url{https://github.com/weeknan/DR-Tune}.
\end{abstract}

\section{Introduction}

Nowadays, it has become a prevailing paradigm to pretrain deep models for common use on large-scale datasets and fine-tune them in multiple diverse downstream tasks in the community of computer vision \cite{he2020mocov1, chen2020simclr}. Due to the data and semantic relevance between pretraining and downstream tasks, the pretrained model implicitly encodes useful prior knowledge, and compared with the ones by training from scratch, it substantially promotes the accuracy of the downstream task and accelerates its training convergence in a variety of applications~\cite{herethinking,transfusion}, \emph{e.g.} image classification, object detection, and semantic segmentation. In particular, when labeled data are quite limited in the downstream task, the issue of over-fitting can be effectively alleviated by using the pretrained model as a training prior.

\begin{figure}
  \centering 
  \includegraphics[width=83mm]{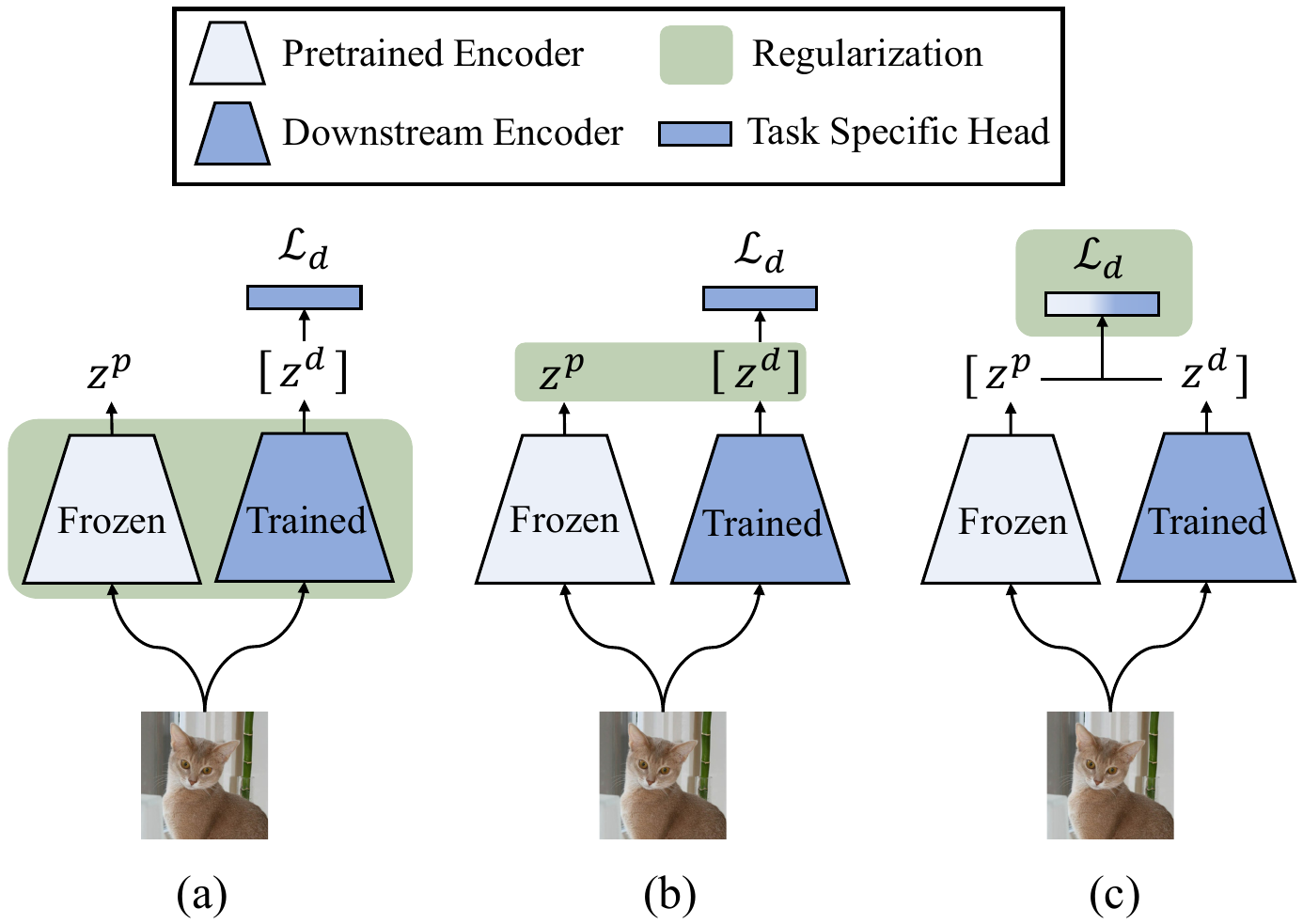}
  \caption{Comparison of distinct regularization-based approaches. (a) (or (b)) performs regularization by reducing the ad-hoc discrepancy between the weights (or the intermediate feature maps) of the downstream encoder and the pretrained one. In contrast, \emph{DR-Tune} (c) performs regularization on the task-specific head by minimizing the classification error with the pretrained feature distribution.}
  \label{reg_compare}
\end{figure}

\begin{figure*}
  \centering 
  \includegraphics[width=170mm]{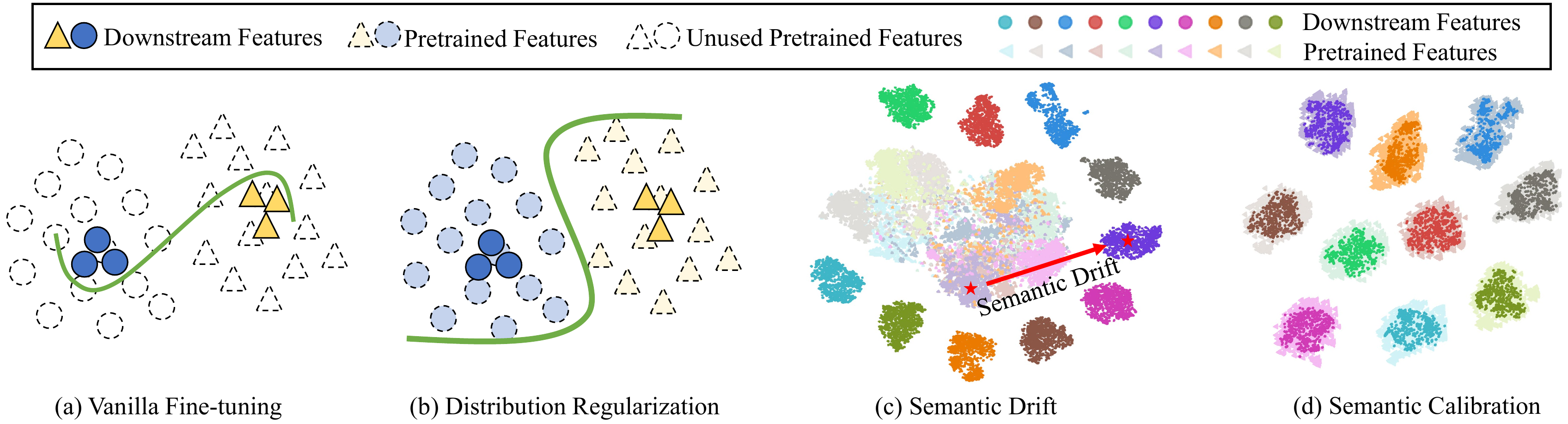}
  \caption{Illustration on the motivation of \emph{DR-Tune}.
  (a) Vanilla fine-tuning only uses downstream features for training, which is prone to be over-fitting. 
  (b) \textbf{Distribution Regularization} employs the pretrained feature distribution to constrain the task head, enforcing it to learn a smooth classification boundary.  
  (c) $t$-SNE \cite{tsne} visualization on the features extracted by the pretrained/downstream encoders on CIFAR10 \cite{cifar}, showing the semantic drift issue. 
  (d) \textbf{Semantic Calibration} clearly alleviates this semantic drift.} 
  \label{motivation}
\end{figure*}

To facilitate training downstream models with the pretrained ones, many efforts have recently been made. One of the typical ways is to directly take the pretrained model for initialization and fine-tune \cite{hinton_finetune, yos_finetune} its weights by elaborately designing task-specific learning objectives \cite{chen2019BSS,li2020rifle,gunel2020scl_tune,zhong2020bi,core}. Nevertheless, these methods neglect retaining the pretrained prior in the fine-tuning phase and tend to incur the ``catastrophic forgetting'' problem \cite{1989catastrophic, recall, distance_reg}, making the learned model prone to over-fit. 

  In contrast, another alternative focuses on utilizing the prior knowledge encoded in the pretrained model to regularize the training of downstream models  \cite{xuhong2018L2SP, distance_reg}. By introducing extra regularization terms based on a pretrained model either on the weights \cite{xuhong2018L2SP} (see Fig.~\ref{reg_compare} (a)) or the intermediate feature maps \cite{komodakis2017paying, li2018delta} (see Fig.~\ref{reg_compare} (b)), these methods prevent the downstream model from over-fitting and significantly boost the overall performance; however, they often impose explicit ad-hoc constraints by reducing the discrepancy between the weights or the sample-wise feature maps generated by the pretrained and downstream models, without considering the semantic drift of the pretrained features. As a consequence, they are inclined to suffer from the non-negligible bias caused by the pretrained model, deteriorating the final result which may be even worse than vanilla fine-tuning in specific scenarios as claimed in \cite{chen2019BSS}, and leave much room for improvement.

 To address the issues above, this paper proposes a novel regularization-based framework for fine-tuning, namely distribution regularization (DR) with semantic calibration (DR-Tune). As Fig.~\ref{reg_compare} (c) illustrates, different from the existing methods, DR-Tune conducts distribution regularization on the downstream classification head, instead of the encoder. The basic idea behind is to minimize the classification error of the downstream task head according to the pretrained feature distribution in addition to the normally used downstream feature distribution. Unfortunately, the discrepancy between the dynamically updated downstream model and the frozen pretrained model incurs semantic drift between the two distributions as shown in Fig.~\ref{motivation} (c), which hinders the task head from learning correct classification boundaries. To alleviate this drift, we develop the semantic calibration (SC) module to align the pretrained and downstream feature distributions via a holistic rotation matrix as well as a group of class-level translation vectors, which are efficiently estimated by establishing two memory banks. The rotation matrix performs global distance-preserving alignment, while the translation vectors offer the alignment of class center pairs, significantly removing the semantic drift as depicted in Fig.~\ref{motivation} (d).

 Intuitively, the proposed DR-Tune framework has two underlying advantages: 1) DR does not impose explicit constraints neither on the weights nor on the intermediate feature maps, largely facilitating optimizing the downstream encoder towards the downstream task; 
 2) SC greatly reduces the semantic drift and the classification bias is thus alleviated when employing the pretrained feature distribution as regularization, leading to improved fine-tuning results; and 3) as in Fig.~\ref{motivation} (b), by leveraging the extra support from the pretrained feature distribution and the downstream features, the task head benefits generating smoother classification boundaries, restricting the over-fitting risk.

The main contributions are summarized as follows: 

1) We propose a novel fine-tuning framework (DR-Tune), which handles over-fitting by regularizing the task-specific head with the pretrained feature distribution. 

2) We design the SC module to address the semantic drift between the pretrained and downstream feature distributions, effectively decreasing the bias introduced by the regularization from the pretrained models.  

3) We conduct extensive evaluation on popular classification datasets and demonstrate that DR-Tune consistently improves the performance as combined with various network structures under different pretraining schemes. 

\section{Related Work}

\subsection{General Model Fine-tuning}
\label{general}
Most existing fine-tuning methods focus on downstream tasks by elaborately designing task-specific learning objectives. SCL \cite{gunel2020scl_tune}, Bi-tuning \cite{zhong2020bi} and Core-tuning \cite{core} incorporate the supervised contrastive loss \cite{khosla2020scl} with the standard cross-entropy (CE) loss, achieving superior performance on classification tasks. M\&M \cite{zhan2018mix} improves semantic segmentation by utilizing limited pixel-wise annotations in the downstream dataset in conjunction with the triplet loss. Besides, BSS \cite{chen2019BSS} observes that small eigenvalues incur degradation compared to vanilla fine-tuning, and thus penalizes on the eigenvalues of the learned representation. RIFLE \cite{li2020rifle} performs fine-tuning by periodically re-initializing the fully connected layers. 
In general, the methods above neglect retaining the pretrained prior in the fine-tuning phase and tend to over-fit on the downstream task.

In addition, several studies also attempt to apply various adapters \cite{residual_adapter, adapter, side-tuning, conv_adapter, ssf} or prompts \cite{jia2022vpt, nie2022pro_tuning, ju2021prompting_vl, bahng2022exploring} to decrease the computational and storage cost during fine-tuning. Despite their efficiency, these methods sacrifice the performance in accuracy.

\subsection{Regularization for Model Fine-tuning} 
\label{reg_method}

Regularization is a prevailing way to make use of the pretrained prior knowledge for fine-tuning. 
Li \etal \cite{xuhong2018L2SP} apply the $\ell^2$-norm penalty between the parameters of the pretrained and downstream models, which outperforms the standard weight decay. 
Yim \etal \cite{yim2017gift} introduce the knowledge distillation \cite{hinton2015distilling, romero2014fitnets} and adopt the distance between the flow of the solution procedure matrix of the pretrained and downstream models as the regularizer. AT \cite{komodakis2017paying} and DELTA \cite{li2018delta} exploit the attention mechanism and regularize the discrepancy between the intermediate feature maps. \cite{distance_reg} assembles multiple distance-based metrics for regularization, which is optimized by the projected gradient descent method. Co-Tuning \cite{you2020co} explores the semantic information of the pretrained dataset and uses the pretrained labels to regularize the fine-tuning process. These methods handle overfitting by imposing explicit ad-hoc constraints to reduce the discrepancy between the weights or sample-wise feature maps of the pretrained and downstream models, but they do not take into account the semantic drift of the pretrained features, thus leaving room for improvement.

Compared to existing solutions as described in Sec. \ref{general} and Sec. \ref{reg_method}, we prevent the downstream model from over-fitting by introducing distribution regularization (DR) on the task head. DR leverages the pretrained feature distribution to enforce the task head learning smooth classification boundaries without imposing explicit constraints on backbones, thus facilitating optimizing the downstream encoder. In addition, we observe the semantic drift between the pretrained and downstream feature distributions, and mitigate it by developing a novel semantic calibration (SC) module, which substantially improves the final performance.

\section{Approach}
\label{sec:Method}

\subsection{Preliminaries}

Suppose a pretrained model $g_{\bm{\phi}^{p}}\cdot f_{\bm{\theta}^p}(\cdot)$, where $f_{\bm{\theta}^p}$ and $g_{\bm{\phi}^{p}}$ denote the encoder and the pretraining task head parameterized by $\bm{\theta}^p$ and $\bm{\phi}^{p}$, respectively. Given a set of training data $\bm{D}=\{(\bm{x}_i^d, y_i) \}^{N}_{i=1}$ for the downstream task, we aim to learn a downstream model  $g_{\bm{\phi}^d}\cdot f_{\bm{\theta}^d}(\cdot)$ by fine-tuning the pretrained model $g_{\bm{\phi}^p}\cdot f_{\bm{\theta}^p}(\cdot)$, where $\bm{x}_i^d$ refers to the $i$-th image with the class label $y_i$, $\bm{\theta}^d$ and $\bm{\phi}^d$ are the parameters to be learned for the downstream encoder $f_{\bm{\theta}^d}$ and the downstream task head $g_{\bm{\phi}^d}$, respectively. 

To learn $\bm{\theta}^d$ and $\bm{\phi}^d$, vanilla fine-tuning firstly applies the pretrained parameter $\bm{\theta}^p$ to initialize $\bm{\theta}^d$ as $\bm{\theta}^d(0):=\bm{\theta}^p$. $\bm{\phi}^d$ is randomly initialized, which is thereafter jointly learned with $\bm{\theta}^d$ by optimizing the following objective:
\begin{equation}
  (\bm{\theta}^d_\ast, \bm{\phi}^d_\ast) = \arg \min\limits_{\bm{\theta}^d, \bm{\phi}^d} \mathcal{L}\left(g_{\bm{\phi}^d}\cdot f_{\bm{\theta}^d};\bm{D}\right), 
  \label{vanilla finetune}
\end{equation}
where $\mathcal{L}(\cdot)$ is the task-specific loss. The fine-tuned model $g_{\bm{\phi}^d_\ast}\cdot f_{\bm{\theta}^d_\ast}$ is used for inference in the downstream task. 

Nevertheless, the vanilla fine-tuning strategy is prone to be over-fitting on the downstream data, especially when the training size $N$ is small. To overcome this shortcoming, the regularization-based fine-tuning strategy is employed by introducing a regularization term $\mathcal{R}(\cdot)$ on $\bm{\theta^{d}}$ according to $\bm{\theta}^{p}$ and optimizing the following objective:
\begin{equation}
  (\bm{\theta}_\ast^d, \bm{\phi}^d_\ast) = \arg \min\limits_{\bm{\theta}^d, \bm{\phi}^d} \mathcal{L}\left(g_{\bm{\phi}^d}\cdot f_{\bm{\theta}^d};\bm{D}\right)+\mathcal{R}\left(\bm{\theta}^{d};\bm{\theta}^{p}\right).
  \label{vanilla finetune_reg}
\end{equation}

\begin{figure*}
  \centering 
  \includegraphics[width=170mm]{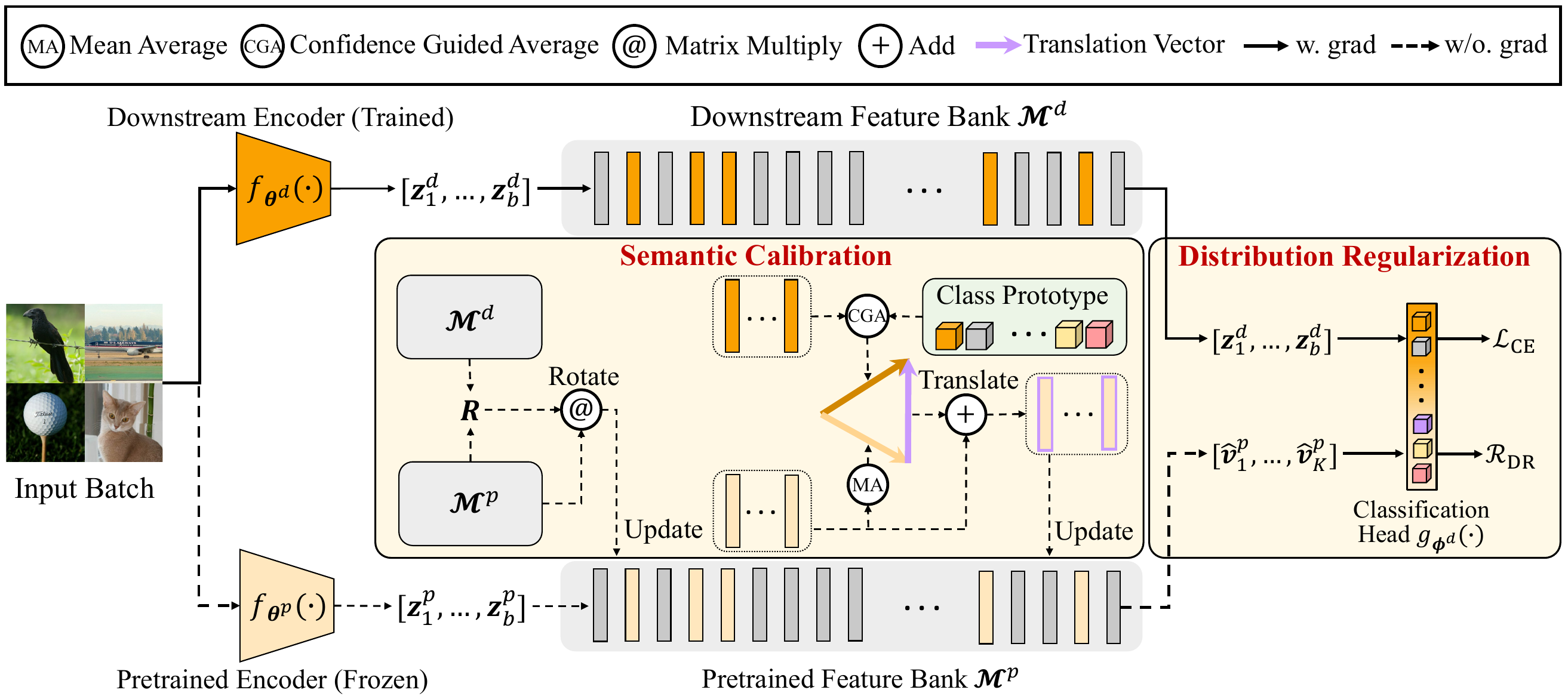}
  \caption{Illustration of the DR-Tune framework. DR-Tune has two branches, 
  including a frozen pretrained encoder $f_{\theta^{p}}$ and a trained downstream encoder $f_{\theta^{d}}$. 
  For input images, we obtain two sets of features extracted by $f_{\theta^{p}}$ and $f_{\theta^{d}}$ respectively 
  and then we store them in their individual feature banks $\bm{\mathcal{M}}^{p}$ and $\bm{\mathcal{M}}^{d}$. 
  Semantic Calibration is further applied to $\bm{\mathcal{M}}^{p}$ to alleviate the semantic drift. Finally, 
  we combine the calibrated pretrained features with the downstream ones to optimize the classification head 
  (\ie Distribution Regularization).
  }
  \label{pipeline}
\end{figure*}

Most of existing fine-tuning methods perform regularization in an ad-hoc manner such as the weight-based ones formulated as $\mathcal{R}=\|\bm{\theta}^{d}-\bm{\theta}^{p}\|$ as well as the feature-based ones written as 
$\mathcal{R}=\sum_{i=1}^{N}\|FM(\bm{x}_{i}^d|f_{\bm{\theta}^{d}})-FM(\bm{x}_{i}^d|f_{\bm{\theta}^{p}})\|$, where $FM(\bm{x}_{i}^d|f_{\bm{\theta}^{d}})$ indicates the feature map of $\bm{x}_{i}^d$ extracted from the intermediate layer of $f_{\bm{\theta}^{d}}$. The former imposes strong constraints on $\bm{\theta}^{d}$, and the later forces the downstream feature $FM(\bm{x}_{i}^d)$ to be the same as the pretrained one for each training sample $\bm{x}_{i}^d$, both of which impede $\bm{\theta}^{d}$ from being sufficiently optimized towards the downstream task.

\subsection{Framework Overview}
\label{subsec:framework}
To address the issues above, we propose a novel fine-tuning framework, namely distribution regularization with semantic calibration (DR-Tune). 

As illustrated in Fig.~\ref{pipeline}, given training set $\bm{D}=\{(\bm{x}_{i}^d, y_{i})\}$, we extract the downstream representations $\{\bm{z}^d_{i}|\bm{z}^d_{i}=f_{\bm{\theta}^{d}}(\bm{x}_{i}^d)\}$ and the pretrained representations $\{\bm{z}^p_{i}|\bm{z}^p_{i}=f_{\bm{\theta}^{p}}(\bm{x}_{i}^d)\}$ by the encoders $f_{\bm{\theta}^{d}}$ and $f_{\bm{\theta}^{p}}$, respectively. 

The basic idea of DR-Tune is employing an implicit distribution regularization (DR) $\mathcal{R}_{\rm DR}(\{(\bm{z}^p_{i},y_{i})\}|g_{\bm{\phi}^{d}})$ on the downstream model, \emph{i.e.} the task head $g_{\bm{\phi}^{d}}$ is enforced to correctly classify the pretrained representations $\{\bm{z}_{i}^p\}$, besides the downstream ones $\{\bm{z}_{i}^d\}$.

However, as shown in Fig.~\ref{motivation} (c), there exists semantic drift between the pretrained feature distribution and the downstream one. Therefore, directly using $\{\bm{z}_{i}^p\}$ for regularization incurs non-negligible bias, thus degrading the performance of the fine-tuned downstream model. To solve this problem, DR-Tune introduces a semantic calibration (SC) module to alleviate the distribution drift. Concretely, as displayed in Fig.~\ref{pipeline}, DR-Tune employs two queues to build a downstream feature bank $\bm{\mathcal{M}}^{d}$ as well as a pretrained feature bank $\bm{\mathcal{M}}^{p}$, which are dynamically updated according to the features $\{\bm{z}_{i}^d\}$ and $\{\bm{z}_{i}^p\}$ in the mini-batch, respectively. $\bm{\mathcal{M}}^{d}$ and $\bm{\mathcal{M}}^{p}$ efficiently represent the downstream and pretrained feature distribution, based on which the calibration parameters including a global rotation matrix $\bm{R}$ and a group of class-level translations $\{\bm{\delta}_{c}\}$ are estimated, where $\bm{\delta}_{c}$ is the translation vector for the $c$-th class. During training, the calibrated pretrained features $\{\hat{\bm{z}}^{p}_{i}|\hat{\bm{z}}^{p}_{i}=\bm{R}\cdot \bm{z}^{p}_{i}+\bm{\delta}_{y_{i}}\}$ are used to form the final distribution regularization as 
$\mathcal{R}_{\rm DR}(\{(\bm{\hat{z}}^p_{i},y_{i})\}|g_{\bm{\phi}^{d}})$. 
In the testing phase, we skip the SC module as well as the feature banks, and only use the downstream encoder $f_{\bm{\theta}^{d}}$ and the head $g_{\bm{\phi}^{d}}$ for inference.

The details about the DR term and the SC module are described in Sec.~\ref{sec:dr} and Sec.~\ref{sec:calibrate}, respectively.

\subsection{Fine-tuning with Distribution Regularization}
\label{sec:dr}
In this section, we elaborate the formulation of DR, \emph{i.e.} 
$\mathcal{R}_{\rm DR}(\{(\bm{z}^p_{i},y_{i})\}|g_{\bm{\phi}^{d}})$.

Formally, suppose the training set $\bm{D}$ is drawn from the data distribution $\mathcal{X}^{d}$, the feature distributions of $\{f_{\bm{\theta}^{d}}(\bm{x}_{i}^{d})\}$ and $\{f_{\bm{\theta}^{p}}(\bm{x}_{i}^{d})\}$ are formulated as  $\mathcal{Z}^{d}=P_{\bm{x} \sim \mathcal{X}^d}(f_{\bm{\theta}^d}(\bm{x}))$ and $\mathcal{Z}^{p}=P_{\bm{x} \sim \mathcal{X}^d}(f_{\bm{\theta}^p}(\bm{x}))$, respectively. 
It is worth noting that both $\mathcal{Z}^{p}$ and $\mathcal{Z}^{d}$ are derived  from the same distribution 
$\mathcal{X}^d$, but by distinct encoders $f_{\bm{\theta}^p}$ and $f_{\bm{\theta}^d}$.

Usually, the downstream task-specific learning objective $\mathcal{L}$ can be briefly written as below:
\begin{equation}
\mathcal{L}=-\log Pr_{\bm{x}_{i}^d\sim \mathcal{X}^{d}}\left(\{(\bm{z}_{i}^d, y_{i})\}|f_{\bm{\theta}^{d}};g_{\bm{\phi}^{d}}\right),
\label{eq:ft_loss}
\end{equation}
where $\bm{z}_{i}^d=f_{\bm{\theta}^{d}}(\bm{x}_{i}^d)$ and $ Pr_{\bm{x}_{i}^d\sim \mathcal{X}^{d}}\left(\{(\bm{z}_{i}^d, y_{i})\}|f_{\bm{\theta}^{d}};g_{\bm{\phi}^{d}}\right)$ is the joint probability of the training feature set $\{(\bm{z}_{i}^d, y_{i})\}$ conditioned on $f_{\bm{\theta}^{d}}$ and $g_{\bm{\phi}^{d}}$.

As aforementioned, $\mathcal{R}_{\rm DR}$ aims to regularize the task head $g_{\bm{\phi}^{d}}$ by enforcing it to classify the pretrained representations $\{\bm{z}^p_{i}\}$. To this end, we adopt the following formulation of $\mathcal{R}_{\rm DR}$
\begin{equation}
\mathcal{R}_{\rm DR}=-\log Pr_{\bm{z}_{i}^p\sim \mathcal{Z}^{p}}\left(\{(\bm{z}_{i}^p, y_{i})\}|g_{\bm{\phi}^{d}}\right),
\label{eq:dr}
\end{equation}
where $y_i$ is the category of $\bm{z}_{i}^p$.
From Eq.~\eqref{eq:dr}, it can be observed that $g_{\bm{\phi}^{d}}$ is optimized to maximize the joint probability of $\{(\bm{z}_{i}^p, y_{i})\}$ when minimizing $\mathcal{R}_{\rm DR}$, thus forcing $g_{\bm{\phi}^{d}}$ to correctly classify $\{\bm{z}_{i}^p\}$.

This kind of regularization has the following advantages compared to existing ad-hoc regularizers: \textbf{1)} $\mathcal{R}_{\rm DR}$ does not impose any explicit constraints neither on the downstream weights $\bm{\theta}^{d}$ nor on the intermediate downstream features, thus bypassing the interference of improper constraints on fine-tuning $f_{\bm{\theta}^{d}}$. \textbf{2)} As shown in Fig.~\ref{motivation}~(b), instead of using the ad-hoc sample-wise regularization, $\mathcal{R}_{\rm DR}$ leverages the pretrained feature distribution $\mathcal{Z}^{p}$ for regularization, which explores holistic information to prevent the downstream task head $g_{\bm{\phi}^{d}}$ from over-fitting. In the meantime, when combining $\mathcal{R}_{\rm DR}$ in Eq.~\eqref{eq:dr} with the task-specific loss $\mathcal{L}$ in Eq.~\eqref{eq:ft_loss}, as $g_{\bm{\phi}^{d}}$ becomes more generalizable, $f_{\bm{\theta}^{d}}$ is improved correspondingly. Please refer to the \emph{supplementary material} for more analysis.
 
To specify the form of $\mathcal{R}_{\rm DR}$, we clarify the joint probability in Eq.~\eqref{eq:dr}. By assuming the independent sampling of $(\bm{z}^{p}_{i},y_{i})$, Eq.~\eqref{eq:dr} is rewritten as $\mathcal{R}_{\rm DR}=-\sum_{\bm{z}_{i}^p\sim \mathcal{Z}^{p}} \log Pr\left((\bm{z}_{i}^p, y_{i})|g_{\bm{\phi}^{d}}\right)$. For the classification task with $C$ classes, the parameters of $g_{\bm{\phi}^{d}}$ can be decomposed as $\bm{\phi}^{d} =[\bm{\phi}_1^{d}, \bm{\phi}_2^{d}, \cdots, \bm{\phi}_C^{d}]$, where $\bm{\phi}_c^{d}$ corresponds to the ones for the $c$-th class prototype. Similar to the CE loss, given a pretrained sample $(\bm{z}^{p}_{i},y_{i})$, the conditional probability  $Pr\left((\bm{z}_{i}^p, y_{i})|g_{\bm{\phi}^{d}}\right)$ turns to be 
$$
Pr\left((\bm{z}_{i}^p, y_{i})|g_{\bm{\phi}^{d}}\right)=\frac{\mathrm{exp}(\bm{\phi}_{y_i} \cdot \bm{z}_i^p)}{\sum_{c = 1}^{C}\mathrm{exp}(\bm{\phi}_c \cdot \bm{z}_i^p)}.
$$

Ideally, all pretrained representations $\{\bm{z}_{i}^p\}$ of the training set should involve in computation of $\mathcal{R}_{\rm DR}$; however it is extremely inefficient to train $g_{\bm{\phi}^{d}}$ by using all of them in each iteration. An alternative way is to extract a mini-batch, but it only captures local information of the distribution. Inspired by  \cite{insdis,he2020mocov1,xbj}, we make a trade-off by employing a feature bank to approximate the distribution $\mathcal{Z}^{p}$. Specifically, we maintain a queue $\bm{\mathcal{M}}^{p}=\{\bm{v}^{p}_{k}\}_{k=1}^{K}$ with a fixed size $K$ by enqueuing the newest features (\emph{i.e.} the features from a mini-batch), and dequeuing the oldest ones. 

Based on $Pr\left((\bm{z}_{i}^p, y_{i})|g_{\bm{\phi}^{d}}\right)$ and $\bm{\mathcal{M}}^{p}$, $\mathcal{R}_{\rm DR}$ is finally formulated as below:
\begin{equation}
  \mathcal{R}_{\rm DR} = - \frac{1}{K} \sum\limits_{k=1}^{K} \log 
  \frac{\mathrm{exp}(\bm{\phi}_{y_k} \cdot \bm{v}_k^p)}{\sum_{c = 1}^{C}\mathrm{exp}(\bm{\phi}_c \cdot \bm{v}_k^p)}.
  \label{extend_ce}
\end{equation}

As to the task-specific loss for fine-tuning, we adopt the commonly used CE loss: 
\begin{equation}
  \mathcal{L}:=\mathcal{L}_{\rm CE} = - \frac{1}{B} \sum_{i = 1}^{B} \log 
  \frac{\mathrm{exp}(\bm{\phi}_{y_i} \cdot f_{\bm{\theta}^{d}}(\bm{x}_i^d))}{\sum_{c = 1}^{C}\mathrm{exp}(\bm{\phi}_c \cdot f_{\bm{\theta}^{d}}(\bm{x}_i^d))}, 
  \label{vanilla_ce}
\end{equation}
where $\{(\bm{x}_{i}^d,y_{i})\}$ is the mini-batch for computational efficiency, and $B$ is the mini-batch size.

\begin{table*}\small  
	\centering 
  \setlength{\tabcolsep}{5.7pt}
	\begin{tabular}{l c c c c c c c c c c} 
      \toprule 
      \makecell[l]{Method} & ImageNet20 & CIFAR10 & CIFAR100 & DTD & Caltech101 & Cars & Pets & Flowers & Aircraft & Avg. \\
      \midrule
      \makecell[l]{CE-tuning }                        & 88.28 & 94.70 & 80.27 & 71.68 & 91.87 & 88.61 & 89.05 & 98.49 & 86.87 & 87.76\\
      \makecell[l]{L2SP \cite{xuhong2018L2SP}}        & 88.49 & 95.14 & 81.43 & 72.18 & 91.98 & 89.00 & 89.43 & 98.66 & 86.55 & 88.10\\
      \makecell[l]{DELTA \cite{li2018delta}}          & 88.35 & 94.76 & 80.39 & 72.23 & 92.19 & 88.73 & 89.54 & 98.65 & 87.05 & 87.99\\		
      \makecell[l]{M\&M \cite{zhan2018mix}}           & 88.53 & 95.02 & 80.58 & 72.43 & 92.91 & 88.90 & 89.60 & 98.57 & 87.45 & 88.22\\	
      \makecell[l]{BSS \cite{chen2019BSS}}            & 88.34 & 94.84 & 80.40 & 72.22 & 91.95 & 88.50 & 89.50 & 98.57 & 87.18 & 87.94\\
      \makecell[l]{RIFLE \cite{li2020rifle}}          & 89.06 & 94.71 & 80.36 & 72.45 & 91.94 & 89.72 & 90.05 & 98.70 & 87.60 & 88.29\\
      \makecell[l]{SCL \cite{gunel2020scl_tune}}      & 89.29 & 95.33 & 81.49 & 72.73 & 92.84 & 89.37 & 89.71 & 98.65 & 87.44 & 88.54\\
      \makecell[l]{Bi-tuning \cite{zhong2020bi}}      & 89.06 & 95.12 & 81.42 & 73.53 & 92.83 & 89.41 & 89.90 & 98.57 & 87.39 & 88.58\\
      \makecell[l]{Core-tuning \cite{core}}            & 92.73 & 97.31 & 84.13 & 75.37 & 93.46 & 90.17 & \textbf{92.36} & 99.18 & 89.48 & 90.47\\
      
      \makecell[l]{SSF* \cite{ssf}}                     & 94.72 & 95.87 & 79.57 & 75.39 & 90.40 & 62.22 & 84.89 & 92.15 & 62.38 & 81.95\\
      \makecell[l]{\textbf{DR-Tune (Ours)}}                 & \textbf{96.03} & \textbf{98.03} & \textbf{85.47} & \textbf{76.65} & \textbf{95.77} &
                                                      \textbf{90.60} & 90.57 & \textbf{99.27} & \textbf{89.80} & \textbf{91.35} \\ 
      \bottomrule 
    \end{tabular}
  \caption{Comparison of the top-1 accuracy (\%) by using various fine-tuning methods based on the self-supervised pretrained model, \emph{i.e.} ResNet-50 pretrained by MoCo-v2 on ImageNet. 
  `*' indicates that the method is re-implemented. The best results are in \textbf{bold}.}
  \label{overall_compare}
\end{table*}

\begin{table*}\small  
	\centering 
  \setlength{\tabcolsep}{10pt}
	\begin{tabular}{l c c c c c c c c} 
      \toprule 
      \makecell[l]{Method} & CIFAR100$^\dagger$ & Caltech101$^\dagger$ & DTD$^\dagger$ & Flowers$^\dagger$ & Pets$^\dagger$ & SVHN & Sun397 & Avg.\\
      \midrule
      \makecell[l]{Linear probing} & 63.4 & 85.0 & 63.2 & 97.0 & 86.3 & 36.6 & 51.0 & 68.93\\
      \makecell[l]{Adapter \cite{adapter_nlp}} & 74.1 & 86.1 & 63.2 & 97.7 & 87.0 & 34.6 & 50.8 & 70.50\\
      \makecell[l]{Bias \cite{bitfit}} & 72.8 & 87.0 & 59.2 & 97.5 & 85.3 & 59.9 & 51.4 & 73.30\\
      \makecell[l]{VPT \cite{jia2022vpt}} & 78.8 & 90.8 & 65.8 & 98.0 & 88.3 & 78.1 & 49.6 & 78.49\\
      \makecell[l]{SSF \cite{ssf}} & 69.0 & 92.6 & \textbf{75.1} & \textbf{99.4} & 91.8 & 90.2 & 52.9 & 81.57\\
      \makecell[l]{Core-tuning* \cite{core}} & 66.3 & 89.7 & 70.9 & 99.0 & 92.3 & 76.4 & 52.5 & 78.16 \\
          \makecell[l]{\textbf{DR-Tune (Ours)}} & \textbf{81.1} & \textbf{92.8} & 71.4 & 99.3 & \textbf{92.4} & \textbf{92.0} & \textbf{54.5} & \textbf{83.36} \\	

      \bottomrule 
    \end{tabular}

  \caption{Comparison of the top-1 accuracy (\%) by using various fine-tuning methods based on the supervised pretrained model, \emph{i.e.} ViT-B pretrained on ImageNet. `*' indicates that the method is re-implemented and `$\dagger$' refers to the training/test split setting as in \cite{vtab}. The best results are in \textbf{bold}.}
  \label{vtab_sup_compare}
\end{table*}

\subsection{Semantic Calibration}
\label{sec:calibrate}

Since the downstream model is dynamically updated during fine-tuning while the pretrained model is kept frozen, the discrepancy between these two models tends to incur a semantic drift between the pretrained feature distribution $\mathcal{Z}^{p}$ and the downstream one $\mathcal{Z}^{d}$ as illustrated in 
Fig.~\ref{motivation} (c). Ignoring this drift and forcing $g_{\bm{\phi}^{d}}$ to classify features from disparate distributions by jointly optimizing $\mathcal{R}_{\rm DR}$ in Eq.~\eqref{extend_ce} and $\mathcal{L}_{CE}$ in Eq.~\eqref{vanilla_ce} degrades the performance.

To alleviate the semantic drift, we attempt to estimate a transformation to calibrate $\mathcal{Z}^{p}$ w.r.t. $\mathcal{Z}^{d}$. To overcome the dilemma in balancing the efficiency and accuracy, we maintain a downstream feature bank $\bm{\mathcal{M}}^{d}=\{\bm{v}^{d}_{k}\}_{k=1}^{K}$ with size $K$, similar to the pretrained one $\bm{\mathcal{M}}^{p}=\{\bm{v}^{p}_{k}\}_{k=1}^{K}$ constructed in the previous section. It is worth noting that $\bm{v}^{d}_{k}$ and $\bm{v}^{p}_{k}$ are two distinct representations for the same image $\bm{x}_{k}$.

In practice, the semantic drift between $\mathcal{Z}^{d}$ and $\mathcal{Z}^{p}$ is extremely complicated, and is hard to estimate. In our work, we simplify it by assuming that the drift is mainly caused by a misalignment of global rotation and a set of local ones of the class centers. Accordingly, we calculate a rotation matrix $\bm{R}$ and the class-level translations $\{\bm{\delta}_{c}\}_{c=1}^{C}$. 

In regards of $\bm{R}$, we estimate it by solving the following optimization problem:
\begin{equation}
    \setlength{\abovedisplayskip}{3pt}
    \setlength{\belowdisplayskip}{3pt}
   \bm{R}={\rm{argmin}}_{\bm{R}'\cdot \bm{R}'^T=\bm{I}_{d}} \sum_{k=1}^{K} \parallel \bm{R}'\cdot \bm{v}_k^p - \bm{v}_k^d \parallel^2, \\
\label{eq:rot}
\end{equation}
where $\bm{I}_{d}$ is a $d$-dimensional identity matrix.

Eq.~\eqref{eq:rot} can be solved by applying SVD on 
the covariance matrix between $\bm{\mathcal{M}}^{p}$ and $\bm{\mathcal{M}}^{d}$ \cite{svd}.

As for the class-level translations $\{\bm{\delta}_{c}\}_{c=1}^{C}$, we observe that the inter-class distribution of $\mathcal{Z}^{p}$ is less discriminative due to the lack of supervision in the downstream task. In contrast, $\mathcal{Z}^{d}$ is more competent at distinguishing different classes. Therefore, we maintain $\mathcal{Z}^{p}$ and use the translation transformation to adjust the inter-class distribution of $\mathcal{Z}^{p}$ to be consistent with $\mathcal{Z}^{d}$. More visualization is given in the \emph{supplementary material}.

With the motivation above, we first estimate the $c$-th class center for $\mathcal{Z}^{p}$ based on $\bm{\mathcal{M}}^{p}$ as below
\begin{equation}
  \bm{\mu}_c^p=\frac{1}{N_c} \sum_{k=1}^{K} \mathbb{I}\left[y_k^p=c\right] \cdot \bm{R} \cdot \bm{v}_k^p.
  \label{eq:pre-train-center}
\end{equation}
In Eq.~\eqref{eq:pre-train-center}, $N_c$ is the number of pretrained features from the $c$-th class, and $\mathbb{I}[y_k=c]$ is the indicator function, which equals to 1 if  $y_k=c$ and 0 otherwise. 

As for the downstream features, we compute the class center based on $\bm{\mathcal{M}}^{d}$ in a more elaborative way as follows
\begin{equation}
  \bm{\mu}_c^d= \sum_{k=1}^{K} \alpha_k \cdot \mathbb{I}\left[y_k^{d}=c\right]\cdot \bm{v}_k^d,
  \label{downstream-center}
\end{equation}
where the weight
\begin{equation}
    \setlength{\abovedisplayskip}{3pt}
    \setlength{\belowdisplayskip}{3pt}
  \alpha_k=\frac{\mathrm{exp}(\bm{\phi}_{{y}_k^{d}} \cdot \bm{v}_k^d)}{\sum_{j = 1}^{K} \mathbb{I}\left[y_j^{d}=y_k^{d}\right] \,\cdot \mathrm{exp}(\bm{\phi}_{y_j^{d}} \cdot \bm{v}_j^d) },
  \label{weight}
\end{equation}
represents the confidence of $\bm{v}_k^d$ that it is correctly classified  to its label by the head $g_{\bm{\phi}^{d}}$. Since an outlier feature is usually hard to classify, its corresponding weight $\alpha_k$ turns to be small, and the effect of outliers on computing the class center is suppressed, resulting in a more precise estimation.

Based on $\{ \bm{\mu}_c^p \}_{c=1}^C$ and $\{ \bm{\mu}_c^d \}_{c=1}^C$, 
the class-level translation vector for the $c$-th class is estimated as below:
\begin{equation}
  \bm{\delta}_c = \bm{\mu}_c^d - \bm{\mu}_c^p, \quad c=1,\cdots,C.
  \label{eq:diff_vect}
\end{equation}

According to the estimated rotation matrix $\bm{R}$ and the class-level translation vector $\{ \bm{\delta}_c \}_{c=1}^C$, the SC module of $\bm{\mathcal{M}}^{p}$ w.r.t. $\bm{\mathcal{M}}^{d}$ is performed in the following: 
\begin{equation}
  \bm{\hat{v}}_k^p = \bm{R}\cdot \bm{v}_k^p + \bm{\delta}_{y_k^p}, \quad k=1,\cdots,K.
  \label{eq:translation}
\end{equation}

\subsection{Optimization}
\label{optimize}
 
According to the SC module in Eq.~\eqref{eq:translation} and Eq.~\eqref{extend_ce}, the final DR is refined as 
\begin{equation}
  \mathcal{R}_{\rm DR} = - \frac{1}{K} \sum\limits_{k=1}^{K} \log 
  \frac{\mathrm{exp}(\bm{\phi}_{y_k} \cdot \bm{\hat{v}}_k^p)}{\sum_{c = 1}^{C}\mathrm{exp}(\bm{\phi}_c \cdot \bm{\hat{v}}_k^p)}.
  \label{dr}
\end{equation}

The overall objective of DR-Tune is formulated as
\begin{equation}
  \min\limits_{\bm{\theta}^d, \bm{\phi}^d} \ \mathcal{L}_{\rm CE} + \lambda\cdot \mathcal{R}_{\rm DR},
  \label{overall_optim}
\end{equation}
where $\mathcal{L}_{\rm CE}$ is from Eq.~\eqref{vanilla_ce}. $\lambda$ is a hyper-parameter balancing the effect of $\mathcal{L}_{\rm CE}$ and $\mathcal{R}_{\rm DR}$, which is set to $\frac{K}{B}$.

\section{Experimental Results}
In this section, we evaluate the performance of DR-Tune by using distinct pretrained models on widely used datasets, compared with the state-of-the-art counterparts.

\subsection{Datasets}
\label{data_details}

We evaluate DR-Tune on widely used datasets, including ImageNet20 \cite{deng2009imagenet, imagenet20}, CIFAR10 \& 100 \cite{cifar}, DTD \cite{DTD}, Caltech101 \cite{cal101}, Stanford Cars \cite{cars}, Oxford Pets \cite{pets} \& Flowers \cite{flowers}, Aircraft \cite{aircraft}, SVHN \cite{svhn} and Sun397 \cite{sun397}. Please refer to the \emph{supplementary material} for more details. 

\subsection{Details}
\label{exp_details}
By following \cite{ericsson2021how_well,core}, we use ResNet-50 \cite{he2016resnet} 
pretrained by MoCo-v2 \cite{chen2020mocov2} and ViT-B \cite{vit} pretrained in a supervised manner on ImageNet \cite{deng2009imagenet} as the backbone in main experiments. Different pretrained strategies and backbones are also evaluated in Sec \ref{more_exp}. The size (\ie $K$) of the memory banks is set as 2,048 by default.

In most of our experiments, we train for 100 epochs by using the SGD optimizer \cite{bottou2010SGD} with a cosine decay scheduler, where the weight decay and momentum are fixed as $1\times10^{-4}$ and 0.9, respectively. We use the linear decay scheduler on ImageNet20 \cite{imagenet20} and the AdamW \cite{loshchilov2018adamw} optimizer to train the ViT \cite{vit} backbone. Since the mini-batch is augmented before the classification head, we set the learning rate of the classification head  $1+\frac{K}{B}$ times that of the backbone. Similar to \cite{kornblith2019Do_better,chen2020simclr,core}, we utilize random cropping and horizontal flipping for data augmentation with an image size of $224\times 224$ during training, and center cropping during test.

\subsection{Comparison with the State-of-the-art}
\label{exp_results}

In the literature, there are mainly two settings for comparison of different methods, \emph{i.e.} the one based on the \emph{self-supervised pretrained model} as in \cite{core} and another based on the \emph{supervised pretrained model} as in \cite{vtab}. As for the self-supervised setting, we compare our method with the following state-of-the-arts: 1) the baseline method denoted as CE-tuning, which simply uses the pretrained model for initialization and is successively trained on downstream data by the standard CE loss; 2) the regularization-based methods including L2SP \cite{xuhong2018L2SP} and DELTA \cite{li2018delta}; 3) other fully fine-tuning methods including M\&M \cite{zhan2018mix}, BSS \cite{chen2019BSS}, RIFLE \cite{li2020rifle}, Bi-tuning \cite{zhong2020bi}, SCL \cite{gunel2020scl_tune} and Core-tuning \cite{core}. As to the supervised setting, the representative parameter efficient methods, including the baseline Linear probing, Adapter \cite{adapter_nlp}, Bias \cite{bitfit}, VPT \cite{jia2022vpt} and SSF \cite{ssf}, are selected. It is worth noting that the datasets as well as the training/test split used in these two settings are \textbf{NOT} the same; therefore we separately report their results for fair comparison as in Table \ref{overall_compare} and Table \ref{vtab_sup_compare}, respectively.

\begin{table} \small
	\centering 
	\resizebox{\linewidth}{!}{
	\begin{tabular}{l c c c c} 	
	\toprule 
	\multirow{2}{*}{\makecell[c]{Pretraining\\Strategy}} & \multicolumn{2}{c}{Caltech101} & \multicolumn{2}{c}{ImageNet20} \\
     ~  & CE-tuning & \textbf{Ours} & CE-tuning & \textbf{Ours} \\ 
	\midrule
    \makecell[l]{MoCo-v1 \cite{he2020mocov1}} & 91.18 & \textbf{91.94} & 86.89 & \textbf{94.83} \\
    \makecell[l]{PCL \cite{pcl}}              & 93.48 & \textbf{94.90} & 83.91 & \textbf{95.80} \\
    \makecell[l]{InfoMin \cite{infomin}}      & 93.38 & \textbf{95.10} & 86.52 & \textbf{96.53} \\
    \makecell[l]{HCSC \cite{hcsc}}            & 93.89 & \textbf{95.73} & 84.10 & \textbf{96.21} \\
    \midrule
    \makecell[l]{SwAV \cite{swav}}            & 92.79 & \textbf{93.94} & 94.62 & \textbf{95.34} \\
    \midrule
    \makecell[l]{SimSiam \cite{simsiam}}      & 82.28 & \textbf{90.33} & 91.33 & \textbf{94.82} \\
		\bottomrule 
	\end{tabular}
	}
  \caption{Top-1 accuracy (\%) of DR-Tune by combining with different pretraining strategies based on ResNet-50, compared to the baseline CE-tuning.}
  \label{diff_algo} 
\end{table}

Under the self-supervised pretraining setting, as summarized in Table~\ref{overall_compare}, vanilla fine-tuning (\ie CE-tuning) performs the worst, indicating the necessity of exploring the pretrained model in downstream tasks, instead of simply using it for initialization. By launching DR on the task head and reducing the semantic drift, DR-Tune largely outperforms the regularization-based methods L2SP and DELTA, promoting their top-1 accuracies averaged by 3.25\% and 3.36\%, respectively. The other counterparts such as Bi-tuning and Core-tuning focus on designing loss functions to boost the learning of downstream models without the pretrained model for training, thus prone to over-fit. In contrast, DR-Tune applies the pretrained features to facilitate the task head learning smooth classification boundaries and achieves better performance on most datasets. For instance, the accuracy of DR-Tune exceeds the second best Core-tuning by 3.30\%/1.34\%/2.31\% on ImageNet20/CIFAR100/Caltech101 respectively, and is 0.88\% higher than Core-tuning on average over all datasets. Under the supervised pretraining setting, as Table~\ref{vtab_sup_compare} shows, our method consistently boosts the averaged top-1 accuracy, promoting the second best method SSF by 1.78\%.

Core-tuning and SSF are the most competitive counterparts only under the self-supervised and supervised setting, respectively, and we further re-implement them and evaluate their performance by using the alternative setting, denoted as SSF$^*$ and Core-tuning$^*$. As displayed, they fail to retain high performance when using different pretrained models, while our method yields decent results in both the settings, clearly showing its generalizability.

\begin{table}[!t] 
  \setlength{\tabcolsep}{7pt}
	\centering 
	\resizebox{0.9\linewidth}{!}{
	\normalsize
	\begin{tabular}{l c c c c} 	
		\toprule
    \multirow{2}{*}{Backbone} & \multicolumn{2}{c}{Caltech101} & \multicolumn{2}{c}{DTD} \\
    ~ & CE-tuning & \textbf{Ours} & CE-tuning & \textbf{Ours} \\
    \midrule
    \makecell[l]{R-50}  & 93.38 & \textbf{95.10} & 68.62 & \textbf{77.97} \\
    \makecell[l]{R-101} & 94.23 & \textbf{95.64} & 70.00 & \textbf{78.41} \\
    \makecell[l]{R-152} & 94.48 & \textbf{96.19} & 70.16 & \textbf{71.44} \\
    \makecell[l]{RX-101}& 94.71 & \textbf{96.39} & 72.18 & \textbf{76.70} \\
    \makecell[l]{RX-152}& 94.85 & \textbf{96.44} & 72.45 & \textbf{78.51} \\
    \midrule
    \makecell[l]{ViT-B} & 94.35 & \textbf{96.03} & 73.72 & \textbf{78.02} \\
    \makecell[l]{ViT-L} & 95.64 & \textbf{97.57} & 73.94 & \textbf{78.83} \\
    \bottomrule
	\end{tabular}
	}
  \caption{Top-1 accuracy (\%) of DR-Tune by combining with distinct backbones, compared to the baseline CE-tuning.}
  \label{diff_arch} 
\end{table}

\subsection{Generalizability}
\label{more_exp}
We further evaluate the generalizability of DR-Tune by combining it with distinct pretraining strategies, backbones as well as the scales of the downstream data.

In regards of \emph{different pretraining strategies}, except for MoCo-v2 used in Table~\ref{overall_compare}, we integrate DR-Tune with the pretrained models based on the ResNet-50 backbone by: 1) the contrastive self-supervised methods including MoCo-v1 \cite{he2020mocov1}, PCL \cite{pcl}, InfoMin \cite{infomin} and HCSC \cite{hcsc}; 2) the clustering based self-supervised method SwAV \cite{swav}; and 3) the prediction based self-supervised method SimSiam \cite{simsiam}. 
As shown in Table~\ref{diff_algo}, DR-Tune consistently delivers significant improvement on Caltech101 and ImageNet20 compared to CE-tuning, in regardless of the pretraining strategy used.

With respect to \emph{distinct backbones}, we adopt the widely used residual networks including ResNet(R)-50/-101/-152 and ResNeXt(RX)-101/-152 \cite{resnext} pretrained by InfoMin \cite{infomin}, as well as the vision transformers including ViT-Base (ViT-B)/-Large (ViT-L) \cite{vit} pretrained by MAE \cite{mae}. As shown in Table~\ref{diff_arch}, DR-Tune obtains gains compared to CE-tuning with distinct backbones. The results on ViT further demonstrate that DR-Tune applies to the Masked Image Modeling  pretraining strategy \cite{beit}.

As for \emph{varying data scales in fine-tuning}, we establish training subsets on ImageNet20 by using three sampling ratios, \ie 10\%, 25\% and 50\%. For each setting, we repeat the experiments for three times with distinct random seeds, and report the mean and standard deviation of the top-1 accuracy. As shown in Table \ref{diff_data}, our method substantially outperforms the compared methods. Especially, when the amount of data is extremely limited (\ie 10\%), the performance of most counterparts sharply drops, observing that the top-1 accuracies of CE-tuning, Bi-tuning, Core-tuning and SSF decrease by \textbf{29.9\%}, \textbf{28.6\%}, \textbf{14.1\%} and \textbf{4.53\%} respectively, compared to the ones using 100\% of the training data. By contrast, DR-Tune performs robustly, with only a \textbf{3.3\%} drop in accuracy. 

\begin{table} \small
  \setlength{\tabcolsep}{2.5pt}
	\centering 
	\begin{tabular}{l c c c c} 	
		\toprule
    \multirow{2}{*}{\makecell[l]{Method}} & \multicolumn{4}{c}{Sampling Ratios on ImageNet20} \\
    ~ & 10\% & 25\% & 50\% & 100\% \\
    \midrule
    \makecell[l]{CE-tuning}  & 58.37$\pm$0.63 & 71.10$\pm$0.28 & 80.79$\pm$0.80 & 88.28 \\
    \makecell[l]{Bi-tuning \cite{zhong2020bi}}  & 60.50$\pm$1.11 & 75.86$\pm$0.74 & 83.19$\pm$0.27 & 89.06\\
    \makecell[l]{Core-tuning \cite{core}}& 78.64$\pm$0.58 & 84.48$\pm$0.34 & 89.09$\pm$0.40 & 92.73 \\
    \makecell[l]{SSF* \cite{ssf}}& 90.17$\pm$0.16 & 92.81$\pm$0.11 & 93.71$\pm$0.19 & 94.70 \\
    \makecell[l]{DR-Tune} & \textbf{92.73$\pm$0.17} & \textbf{94.16$\pm$0.20} & \textbf{95.21$\pm$0.07} & \textbf{96.03} \\
    \bottomrule
	\end{tabular}
  \caption{Comparison of the top-1 accuracy (\%) using varying data scales for fine-tuning. `*' indicates our implementation.}
  \label{diff_data} 
\end{table}

\subsection{Ablation Studies}

\textbf{Effect of main components.} We investigate the influences of DR and SC in DR-Tune on the Caltech101, Cars and Pets datasets. All the results are obtained based on ResNet-50 pretrained by MoCo-v2 on ImageNet. As displayed in Table \ref{ablation_compon}, both DR and SC contribute to the overall performance. For fine-grained Cars and Flowers, the feature distributions generated by the pretrained model and the downstream one exhibit a severe semantic drift, due to their large discrepancy on the semantic granularity. DR alone fails to deal with this drift, thus incurring degradation in performance. SC remarkably boosts the overall performance by mitigating this semantic drift. Please refer to the \emph{supplementary material} for more analysis.

\textbf{Effect of different transformations in SC.}
The proposed SC module performs feature transformation by a global rotation (GR) and a group of class-level translations (CLT) refined by the confidence guided average (CGA). We therefore evaluate their effects on Caltech101, Cars and Pets. As demonstrated in Table \ref{ablation_sc}, both GR and CLT clearly promote the performance. By suppressing the weights of suspicious outlier features, CGA facilitates computing the centers more precisely, further improving the accuracy, especially on the fine-grained Cars and Pets datasets, of which the centers are more sensitive to hard samples due to small inter-class discrepancies.

\begin{table} \small
  \setlength{\tabcolsep}{6pt}
	\centering 
	\begin{tabular}{c c c c c c c c} 	
		\toprule
    CE & DR & SC & Caltech101 & Cars & Pets\\
    \midrule
    \checkmark & ~          & ~           & 91.93 & 88.45 & 88.36 \\
    \checkmark & \checkmark & ~           & 94.39 & 89.03 & 89.37 \\
    \checkmark & \checkmark & \checkmark  & \textbf{95.73} & \textbf{90.60} & \textbf{90.57} \\
    \bottomrule
	\end{tabular}
  \caption{Ablation studies on the main components. CE: Cross Entropy; DR: Distribution Regularization; and SC: Semantic Calibration.}
  \label{ablation_compon} 
\end{table}

\begin{table} \small
  \setlength{\tabcolsep}{6pt}
	\centering 
	\begin{tabular}{c c c c c c} 	
		\toprule
    GR & CLT & CGA & Caltech101 & Cars & Pets\\
    \midrule
    ~           & ~         & ~          & 94.39  & 89.03 & 89.37     \\
    \checkmark   & ~         & ~          & 95.59 & 90.25 & 89.62     \\
      ~         & \checkmark & ~          & 95.11 & 89.96 & 89.69 \\
      ~         & \checkmark & \checkmark & 95.17 & 90.29 & 90.24 \\
    \checkmark  & \checkmark & \checkmark & \textbf{95.73} & \textbf{90.60} & \textbf{90.57} \\
    \bottomrule
	\end{tabular}
  \caption{Ablation studies for different operations in SC. GR: Global Rotation; CLT: Class-Level Translation; and CGA: Confidence Guided Average. }
  \label{ablation_sc} 
\end{table}

\begin{figure}
  \centering 
  \includegraphics[width=80mm]{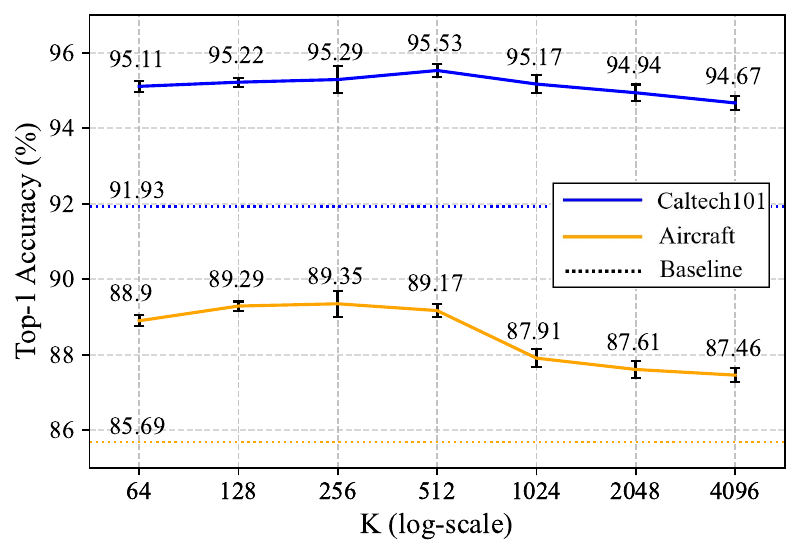}
  \caption{Ablation results on Caltech101 and Aircraft \wrt $K$.} 
  \label{diff-k}
\end{figure}

\textbf{Effect of hyper-parameter.}
The DR-Tune framework is hyper-parameter-friendly, and the only hyper-parameter is the size of the feature banks $K$. Since the learning rate varies as $K$ changes (see details in Sec. \ref{exp_details}) in our setting, we fix it as 0.01 to eliminate its interference. As shown in Fig. \ref{diff-k}, DR-Tune outperforms the baseline by vanilla fine-tuning and performs steadily with different $K$ values, even when $K$ is set at a small one (\emph{e.g.} 64).

\section{Conclusion and Limitation} 
In this paper, we propose a novel framework, namely distribution regularization with semantic calibration (DR-Tune), for fine-tuning pretrained visual models on downstream tasks. DR-Tune employs DR on the classification head by leveraging the pretrained feature distribution, and develops an SC module to alleviate the semantic drift of the pretrained features relative to the downstream ones. Extensive comparison results as well as ablation studies on widely used datasets clearly show the effectiveness and generalizability of the proposed method. 

Despite its merits, DR-Tune has some limitations: 1) It suffers from a high training
latency, due to computation of rotations by SVD in SC,
which can be further improved by more efficient solutions. 2) SC aligns the downstream and pretrained features by a global feature after average pooling for classification, ignoring spatial misalignment, which is crucial to spatio-sensitive tasks, \emph{e.g.} object detection and semantic segmentation, leaving room for gains.

\section*{Acknowledgment} 
This work is partly supported by the National Key R\&D Program of China (2021ZD0110503), the National Natural Science Foundation of China (62022011 and 62202034),  the Research Program of State Key Laboratory of Software Development Environment, and the Fundamental Research Funds for the Central Universities.

{\small
\bibliographystyle{ieee_fullname}
\bibliography{egbib}
}

\appendix
\clearpage

\begin{center}{\bf {\Large Supplementary Material}}
\end{center}

\begin{table*}[!t] \small
  \setlength{\tabcolsep}{5.5pt}
	\centering 
	\begin{tabular}{l  c  c  c  c  c  c  c  c  c} 	
	\toprule
    \makecell[l]{Hyper-parameter} & ImageNet20 & CIFAR10 & CIFAR100 & DTD & Caltech101 & Cars & Pets & Flowers & Aircraft \\
    \midrule
    \makecell[l]{Epochs} & \multicolumn{7}{c}{100} & 200 & 100\\
    \makecell[l]{lr schedule} & linear decay &\multicolumn{8}{c}{cosine decay} \\
    \makecell[l]{lr for the encoder} & 0.01 & 0.01 & 0.01 & 0.01 & 0.1 & 0.1 & 0.01 & 0.01 & 0.1 \\
    \makecell[l]{lr for the head} & 0.33 & 0.33 & 0.33 & 0.33 & 0.1 & 0.1 & 0.17 & 0.13 & 0.1 \\
    \makecell[l]{The size $K$ of memory banks} & 2048 & 2048 & 2048 & 2048 & 2048 & 2304 & 1024 & 768 & 2048 \\
    \makecell[l]{The batch size $B$} & \multicolumn{9}{c}{64} \\
    \makecell[l]{Weight decay factor} & \multicolumn{9}{c}{$10^{-4}$} \\
    \makecell[l]{Momentum factor} & \multicolumn{9}{c}{0.9} \\
    \bottomrule
	\end{tabular}

  \caption{Details about the hyper-parameters used for comparison with the fine-tuning methods based on the  \emph{self-supervised pretrained model}, corresponding to Table 1 of the main body. `lr' is the abbreviation of `learning rate'.}
  \label{hyper-a} 
\end{table*}

\begin{table*}[!t] \small
  \setlength{\tabcolsep}{10pt}
	\centering 
	\begin{tabular}{l  c  c  c  c  c  c  c} 	
	\toprule
    \makecell[l]{Hyper-parameter} & CIFAR100$^\dagger$ & Caltech101$^\dagger$ & DTD$^\dagger$ & Flowers$^\dagger$ & Pets$^\dagger$ & SVHN & Sun397 \\
    \midrule
    \makecell[l]{Epochs} & 100 & 300 &  \multicolumn{5}{c}{100}\\
    \makecell[l]{lr schedule} &\multicolumn{7}{c}{cosine decay} \\
    \makecell[l]{lr for the encoder} & 0.01 & 0.01 & 0.01 & 0.01 & 0.01 & 0.01 & 0.01\\
    \makecell[l]{lr for the head}    & 0.17 & 0.02 & 0.1  & 0.33 & 0.1  & 0.1  & 0.1 \\
    \makecell[l]{The size $K$ of memory banks} & 512 & 128 & 32 & 2048 & 256 & 128 & 2048\\
    \makecell[l]{The batch size $B$} & \multicolumn{7}{c}{32} \\
    \makecell[l]{Weight decay factor} & $10^{-4}$ & $10^{-3}$ & $10^{-3}$ & $10^{-4}$ & $10^{-4}$ & $10^{-3}$ & $10^{-4}$\\
    \makecell[l]{Momentum factor} & \multicolumn{7}{c}{0.9} \\
    \bottomrule
	\end{tabular}
  \caption{Details about the hyper-parameters used for comparison with the fine-tuning methods based on the \emph{supervised pretrained model}, corresponding to Table 2 of the main body. `lr' is the abbreviation of `learning rate'. `$\dagger$' refers to the training/test split setting as in \cite{vtab}.}
  \label{hyper-b} 
\end{table*}

In this document, we describe more details about the datasets and the settings of hyper-parameters used for evaluation in Sec.~\ref{detail}. Additionally, we summarize the overall pipeline of the proposed DR-Tune framework in Sec.~\ref{alg}, and provide more analysis, semantic segmentation results as well as quantitative results in Sec.~\ref{Analysis}, Sec.~\ref{segmentation} and Sec.~\ref{qual}, respectively. Finally, we discuss the limitations in Sec.~\ref{limit}.

\section{Details on Datasets and Hyper-parameters.}
\label{detail}
In Sec.~4 of the main body, we briefly summarize the datasets used for evaluation, including \textbf{ImageNet20} \cite{deng2009imagenet, imagenet20}, \textbf{CIFAR10 \& 100} \cite{cifar}, \textbf{DTD} \cite{DTD}, \textbf{Caltech101} \cite{cal101}, Stanford \textbf{Cars} \cite{cars}, Oxford-IIIT \textbf{Pets} \cite{pets}, Oxford 102 \textbf{Flowers} \cite{flowers}, FGVC \textbf{Aircraft} \cite{aircraft}, \textbf{SVHN} \cite{svhn} and \textbf{Sun397} \cite{sun397}. As a supplement, we describe more details in this section.

\textbf{ImageNet20} is a subset of the large-scale ImageNet dataset \cite{deng2009imagenet}, which contains 26,348 images from 20 categories. It is collected by combining an easy-to-classify dataset Imagenette and a hard-to-classify dataset Imagewoof \cite{imagenet20}. On this dataset, 18,494 images are used for training and the rest 7,854 images are utilized for evaluation. 

\textbf{CIFAR10 \& 100} \cite{cifar} are two widely used datasets containing natural objects from 10 and 100 categories, respectively. They are both divided into a subset of 50,000 images for training and a subset of 10,000 images for evaluation. 

\textbf{Describable Textures Dataset (DTD}) \cite{DTD} is a texture dataset, consisting of 5,640 images organized according to a list of 47 categories inspired from human perception. 3,760 images are used for training and the remaining 1,880 images are adopted for evaluation. 

The \textbf{Caltech101} dataset \cite{cal101} includes 9,146 images from 101 distinct categories, each of which contains 40 to 800 images. We use 3,060 images and 6,084 images for training and evaluation, respectively. 

Stanford \textbf{Cars} \cite{cars} is a fine-grained dataset, which contains 16,185 images of 196 different types of cars. This dataset is split into a set of 8,144 images for training and a set of 8,041 images for evaluation. 

\begin{algorithm}[!tbp]
  \caption{The overall pipeline of DR-Tune.}   
  \label{alg:algo}
  \LinesNumbered
  \KwIn{The pretrained encoder $f_{\bm{\theta}^p}$, the size of the memory bank $K$ and the batch size $B$.}
  \KwOut{The fine-tuned downstream encoder $f_{\bm{\theta}^d}$ and the classification head $g_{\bm{\phi}^d}$.}
  \textbf{Initialization:} Set $\bm{\theta}^d:=\bm{\theta}^p$, randomly initialize $\bm{\phi}^d$, and fill the memory banks $\bm{\mathcal{M}}^{p}$ and $\bm{\mathcal{M}}^{d}$ with the pretrained features. \\
  \While{not converge}{
      Sample a mini-batch $\{ \bm{x}_{i}^d, y_i \}_{i=1}^B$.
      \\
      \For{$i\in\{1,\cdots,B\}$}{
          Extract the pretrained and downstream features for $\bm{x}_{i}^d$ as follows: $\bm{z}^p_{i}=f_{\bm{\theta}^p}(\bm{x}_{i}^d), \bm{z}^d_{i}=f_{\bm{\theta}^d}(\bm{x}_{i}^d)$. \\
        }

        Calculate the rotation matrix $\bm{R}$ via SVD \cite{svd}. \\ 
        Compute the class-level translations as below:\\
        \For{$c=1$ \rm\textbf{to} $C$}{
          Calculate $\bm{\mu}_c^p$ based on $\bm{\mathcal{M}}^{p}$ by Eq.~(8).\\ 
          Calculate $\bm{\mu}_c^d$ based on $\bm{\mathcal{M}}^{d}$ by Eqs.~(9)-(10).\\ 
          Compute the $c$-th translation vector as below\\ $\bm{\delta}_c=\bm{\mu}_c^d - \bm{\mu}_c^p$. \\
        }
        Calibrate the memory bank $\bm{\mathcal{M}}^{p}$ via Eq.~(12).\\ 
        Update $\bm{\theta}^d$ and $\bm{\phi}^d$ by optimizing Eq.~(14).\\ 
    
        Update $\bm{\mathcal{M}}^{p}$/$\bm{\mathcal{M}}^{d}$ by $\bm{z}^p_{i}$/$\bm{z}^d_{i}$, respectively.
    } 
\end{algorithm}

Oxford-IIIT \textbf{Pets} \cite{pets} consists of the images captured from 37 kinds of pets, of which each class roughly includes 200 images. This dataset exhibits large variations in scale, pose and lighting. We use 3,680 images for training and the rest 3,369 images for evaluation. 

Oxford 102 \textbf{Flowers} \cite{flowers} contains 7,370 flower images from 102 different categories. 6,552 images are used for training and 818 images for evaluation. 

The FGVC \textbf{Aircraft} \cite{aircraft} is a fine-grained dataset, which contains 10,000 images from 100 different types of aircraft models. We split this dataset into a subset of 6,667 images for training and the remaining 3,333 images for evaluation. 

\textbf{SVHN} is obtained from house numbers in Google Street View images, including 73,257 training images and 26,032 test images of size 32x32 from 10 classes. By following the training/test split setting as in \cite{vtab}, we adopt 1,000 images for training and 26,032 images for evaluation.

\textbf{Sun397} \cite{sun397} is a scene understanding benchmark with 76,128 training images and 21,750 test images of 397 categories. Following the training/test split setting as in \cite{vtab}, we adopt 1,000 images for training and 21,750 images for evaluation.

\textbf{Settings of hyper-parameters.} As depicted in Sec. 4.3 of the main body, we compare DR-Tune with the state-of-the-art under two different settings, \emph{i.e.} the one based on the self-supervised pretrained model and the other based on the supervised pretrained model. The corresponding settings of hyper-parameters are summarized in Table.~\ref{hyper-a} and Table.~\ref{hyper-b}, respectively.

\section{Overall Pipeline of DR-Tune}
\label{alg}
In Sec. 3 of the main body, we elaborate the technical details on the main components of Dr-Tune. We additionally summarize the overall pipeline of DR-Tune in Algorithm \ref{alg:algo}.

\section{More Analysis on DR-Tune}
\label{Analysis}

In this section, we conduct a more detailed study on how DR-Tune contributes to the performance gain by analyzing the encoder as well as the classification head on the CIFAR-10 benchmark. 
We also analyze some detailed designs in the SC module and compare DR-Tune with knowledge distillation (KD). 
Furthermore, we report the runtime cost and standard errors.

\begin{figure*}[!t]
  \centering 
  \includegraphics[width=170mm]{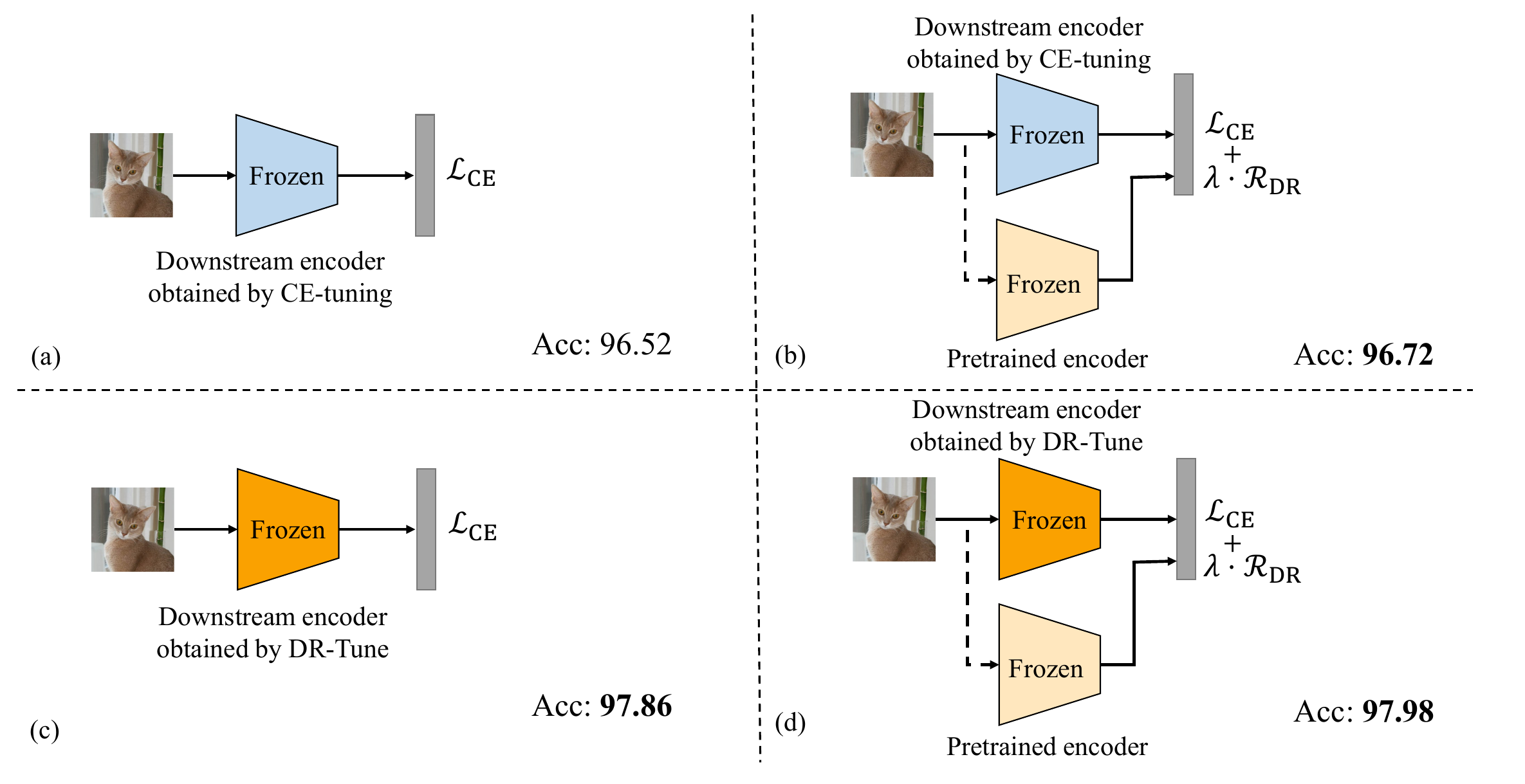}
  \caption{Illustration of different learning strategies: (a) The baseline CE-Tuning; (b) Training the classification head by optimizing $\mathcal{L}_{\rm CE} + \lambda\cdot \mathcal{R}_{\rm DR}$; (c) Applying the downstream encoder generated by DR-Tune; (d) Combining the settings in (b) and (c).}
  \label{exp}
\end{figure*}

\textbf{On the classification head.} In this case, we take a counterpart, which is composed of a frozen downstream encoder fine-tuned by CE-tuning and a classification head randomly initialized. As shown in Fig.~\ref{exp} (a) and (b), the classification head is trained by the standard Cross-Entropy loss (\ie $\mathcal{L}_{\rm CE}$) and the one used in DR-Tune (\ie $\mathcal{L}_{\rm CE} + \lambda\cdot \mathcal{R}_{\rm DR}$), respectively; and we can observe that the top-1 accuracy is improved from 96.52\% to 96.72\%, indicating that $\mathcal{R}_{\rm DR}$ leads to a better classification head.

\textbf{On the encoder.} We compare two models that are depicted in Fig.~\ref{exp} (a) and (c), both of which have a frozen downstream encoder and a randomly initialized classification head and are trained by $\mathcal{L}_{\rm CE}$. Their difference lies in that the downstream encoder is fine-tuned by CE-tuning or by DR-Tune, and this change improves the top-1 accuracy from 96.52\% to 97.86\%, showing that DR-Tune facilitates the training of a stronger encoder.

As shown in Fig.~\ref{exp} (d), when we combine the settings in Fig.~\ref{exp} (b) and (c), the improved encoder and classification head finally reach the top-1 accuracy of 97.98\%, highlighting the effectiveness of DR-Tune.

\begin{table}[t] \small
\setlength{\tabcolsep}{5pt}
\centering 
    \begin{tabular}{l c c c c} 	
    \toprule
    \makecell[l]{Operation} & Imagenet20 & CIFAR10 & Pets &\\
    \midrule
    \makecell[l]{CLR} & 95.82 & 97.75 & 90.19\\
    \makecell[l]{SA} & 95.77 & 97.79 & 90.24\\
    \makecell[l]{GR (w/o SA)} & 95.85 & 97.82 & 89.56\\
    \makecell[l]{GR (\textbf{Ours})} & \textbf{96.03} & \textbf{98.03} & \textbf{90.57}\\
    
    \bottomrule 
    \end{tabular}
    \caption{Top-1 accuracies (\%) of different operations in the SC module.}
  \label{diff_op} 
\end{table}

\begin{table}[t] \small
\setlength{\tabcolsep}{3.5pt}
\centering 

    \begin{tabular}{l c c c c} 	
    \toprule
    \makecell[l]{Method} & Reference & \makecell[c]{Teacher} & Caltech101 & DTD \\
    \midrule
    \makecell[l]{CE-tuning} & - & \makecell[c]{-} & 93.38 & 68.62\\
    \multirow{3}{*}{\makecell[l]{KD \cite{hinton2015distilling}}} & \multirow{3}{*}{NeurIPS'14} & ResNet-50$^\dagger$ & 94.46 & 72.66\\
                            &  & ~~ResNet-101$^\dagger$ & 93.68 & 74.42\\
                            &  & ~~ResNet-101$^*$ & 95.04 & 76.86\\    
    \makecell[l]{RKD \cite{rkd}} & CVPR'19 & ResNet-50$^\dagger$ & 93.66 & 69.10\\
    \makecell[l]{MLD \cite{mld}} & CVPR'23 & ResNet-50$^\dagger$ & 94.90 & 72.82\\
    \makecell[l]{DR-Tune} & \textbf{Ours} & \makecell[c]{-} & \textbf{95.10} & \textbf{77.97}\\
    \bottomrule 
    \end{tabular}
    \caption{Top-1 accuracies (\%) of KD and DR-Tune with ResNet-50 as student network. $^\dagger$: pretrained by InfoMin; $^*$: supervised pretraining.}
  \label{kd} 
\end{table}

\begin{table*}[t] \small
\setlength{\tabcolsep}{3.0pt}
\centering 
    \begin{tabular}{l c c c c c} 	
    \toprule
    \multirow{2}{*}{\makecell[l]{Method}} & \multicolumn{2}{c}{Train} & \multicolumn{3}{c}{Test}\\
    \cmidrule(r){2-3} \cmidrule(r){4-6}
                            & \makecell{Latency$\downarrow$ \\(ms)}  & \makecell{Memory$\downarrow$ \\(GB)} & \makecell{Latency$\downarrow$ \\(ms)}  & \makecell{Memory$\downarrow$ \\(GB)} & \makecell{Accuracy$\uparrow$ \\(\%)}\\
    \midrule
    \makecell[l]{CE-tuning} & 73.55 & 7.64 & 66.68 & 4.22 & 87.76 \\
    \makecell[l]{Core-tuning \cite{core}} & 151.92 & 22.22 & 67.04 & 4.22 & 90.47 \\
    \makecell[l]{DR-Tune (\textbf{Ours})} & 167.50 & 8.41 & 66.49 & 4.22 & 91.35\\
    \bottomrule 
    \end{tabular}
    \caption{Comparison of runtime cost and accuracy.}
  \label{runtime} 
\end{table*}

\textbf{On the SC module.} Global rotation (GR) is performed in the SC module to alleviate the semantic drift. We explore some different designs for this. 
(1) Rotation is performed around the category center of each class, \emph{i.e.} class-level rotation (CLR). (2) Replace the rotation operation by aligning the L2-norm between pretrained and downstream features, \emph{i.e.} scale alignment (SA). As shown in Table \ref{diff_op}, CLR does not lead to a gain, but takes $C-1$ times more operations than GR ($C$: number of classes). We thus adopt GR in implementation. 
The performance of SA is not as good as GR in most cases, but using SA with GR can boost the performance, indicating that using both rotation and scale alignment is a better option.

\textbf{Comparison to knowledge distillation.} The Knowledge distillation (KD) based methods utilize a frozen pretrained teacher network to guide the student network, which has a similar framework with DR-Tune. We thus compare DR-Tune to some representative KD-based methods: 1) logit distillation including KD \cite{hinton2015distilling} and MLD \cite{mld} and 2) feature distillation \emph{i.e.} RKD \cite{rkd}. Despite sharing the same spirit of using pretrained models as regularizers, the KD-based methods ignore the semantic drift issue and impose constraints on the whole downstream model instead of the task head, which may degrade the performance. As an empirical study, Table~\ref{kd} shows that all the KD-based methods boost the accuracy of the baseline CE-tuning, but perform worse than DR-Tune when using the same teacher ResNet-50 pretrained by InfoMin. We then evaluate KD using different teachers with various backbones and pretraining schemes. As displayed, larger teacher models deliver further improvements to KD, but the results are still not as good as those of DR-Tune.

\begin{figure*}[!t]
  \centering 
  \includegraphics[width=170mm]{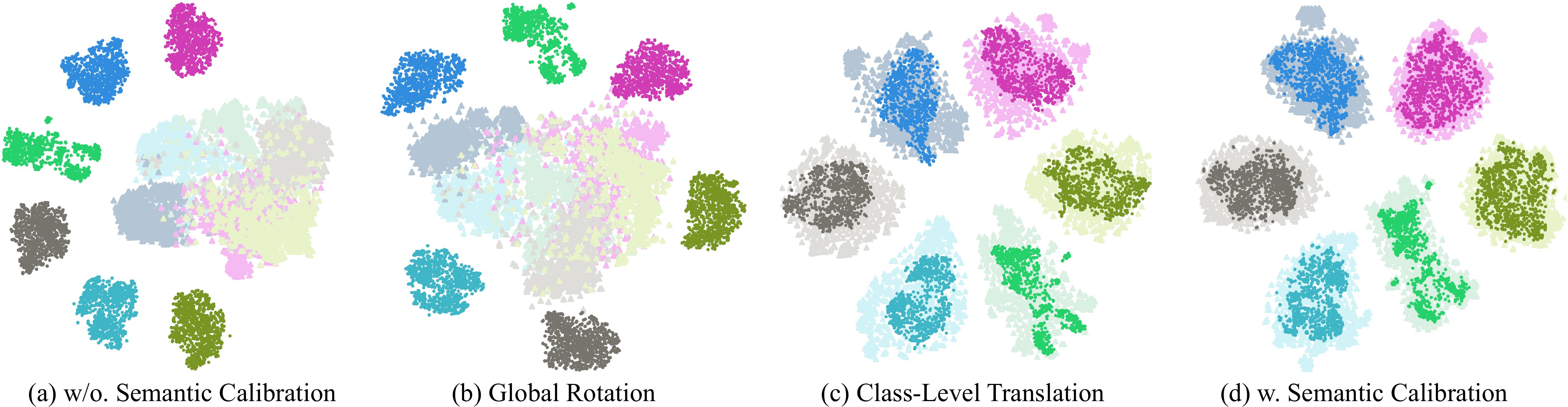}
  \caption{$t$-SNE \cite{tsne} visualization of the pretrained and downstream features on 
  CIFAR10 from the first 6 classes. Different colors indicate different classes, and points with low/high brightness denote the pretrained/downstream features, respectively.}
  \label{visual_sc}
\end{figure*}

\textbf{On the runtime cost.} We report the latency and memory for CE-tuning, Core-tuning and DR-Tune, evaluated using the same NVIDIA V100 GPU with a batch size of 64, based on ResNet-50 pretrained by MoCo-v2. As in Table~\ref{runtime}, DR-Tune has relatively higher training latency compared to CE-tuning, due to extra computation in DR and SC. Core-tuning suffers much more memory usage, as it employs extra parameters and the feature mixture strategy. However, DR-Tune takes a similar cost to CE-tuning in testing, since DR and SC are not used in this phase. Besides, DR-tune delivers remarkably higher accuracies, thus reaching a better balance between efficiency and accuracy for deployment.

\textbf{On the standard errors.} In Table 1 and Table 2 of the main body, we report the mean results after repeating the experiments for three times with different random seeds on each dataset, omitting the standard errors for succinctness. In this supplement, we provide the standard errors to validate the robustness. Note that the counterparts including Linear probing, Adapter, Bias, VPT and SSF in Table 2 do NOT report the standard errors. Therefore, we only report the standard errors of DR-Tune and the re-implemented baseline Core-tuning. The results are summarized in Table \ref{std} and Table \ref{std_2}, showing that our method steadily reaches moderately small standard errors on different datasets and settings.

\begin{table*}[t]\small  
	\centering 
  \setlength{\tabcolsep}{12pt}
	\begin{tabular}{l c c c c c} 
      \toprule 
      \makecell[l]{Method} & ImageNet20 & CIFAR10 & CIFAR100 & DTD & Caltech101\\
      \midrule
      \makecell[l]{CE-tuning }                         & 88.28$\pm$0.47 & 94.70$\pm$0.39 & 80.27$\pm$0.60 & 71.68$\pm$0.53 & 91.87$\pm$0.18 \\
      \makecell[l]{L2SP \cite{xuhong2018L2SP}}        & 88.49$\pm$0.40 & 95.14$\pm$0.22 & 81.43$\pm$0.22 & 72.18$\pm$0.61 & 91.98$\pm$0.07 \\
      \makecell[l]{DELTA \cite{li2018delta}}          & 88.35$\pm$0.41 & 94.76$\pm$0.05 & 80.39$\pm$0.41 & 72.23$\pm$0.23 & 92.19$\pm$0.45 \\		
      \makecell[l]{M\&M \cite{zhan2018mix}}           & 88.53$\pm$0.21 & 95.02$\pm$0.07 & 80.58$\pm$0.19 & 72.43$\pm$0.43 & 92.91$\pm$0.08 \\	
      \makecell[l]{BSS \cite{chen2019BSS}}            & 88.34$\pm$0.62 & 94.84$\pm$0.21 & 80.40$\pm$0.30 & 72.22$\pm$0.17 & 91.95$\pm$0.12 \\
      \makecell[l]{RIFLE \cite{li2020rifle}}          & 89.06$\pm$0.28 & 94.71$\pm$0.13 & 80.36$\pm$0.07 & 72.45$\pm$0.30 & 91.94$\pm$0.23 \\
      \makecell[l]{SCL \cite{gunel2020scl_tune}}      & 89.29$\pm$0.07 & 95.33$\pm$0.09 & 81.49$\pm$0.27 & 72.73$\pm$0.31 & 92.84$\pm$0.03 \\
      \makecell[l]{Bi-tuning \cite{zhong2020bi}}      & 89.06$\pm$0.08 & 95.12$\pm$0.15 & 81.42$\pm$0.01 & 73.53$\pm$0.37 & 92.83$\pm$0.06 \\
      \makecell[l]{Core-tuning \cite{core}}            & 92.73$\pm$0.17 & 97.31$\pm$0.10 & 84.13$\pm$0.27 & 75.37$\pm$0.37 & 93.46$\pm$0.06 \\
      \makecell[l]{SSF* \cite{ssf}}                     & 94.72$\pm$0.07 & 95.87$\pm$0.10 & 79.57$\pm$0.02 & 75.39$\pm$0.66 & 90.40$\pm$0.17 \\
      \makecell[l]{\textbf{DR-Tune (Ours)}}            & \textbf{96.03}$\pm$0.11 & \textbf{98.03}$\pm$0.04 & \textbf{85.47}$\pm$0.08 & \textbf{76.65}$\pm$0.07 & \textbf{95.77}$\pm$0.12\\ 
      \midrule
      \midrule
      \makecell[l]{Method} & Cars & Pets & Flowers & Aircraft & Avg. \\
      \midrule
      \makecell[l]{CE-tuning }                         & 88.61$\pm$0.43 & 89.05$\pm$0.01 & 98.49$\pm$0.06 & 86.87$\pm$0.18 & 87.76\\
      \makecell[l]{L2SP \cite{xuhong2018L2SP}}        & 89.00$\pm$0.23 & 89.43$\pm$0.27 & 98.66$\pm$0.20 & 86.55$\pm$0.30 & 88.10\\
      \makecell[l]{DELTA \cite{li2018delta}}          & 88.73$\pm$0.05 & 89.54$\pm$0.48 & 98.65$\pm$0.17 & 87.05$\pm$0.37 & 87.99\\		
      \makecell[l]{M\&M \cite{zhan2018mix}}           & 88.90$\pm$0.70 & 89.60$\pm$0.09 & 98.57$\pm$0.15 & 87.45$\pm$0.28 & 88.22\\	
      \makecell[l]{BSS \cite{chen2019BSS}}            & 88.50$\pm$0.02 & 89.50$\pm$0.42 & 98.57$\pm$0.15 & 87.18$\pm$0.71 & 87.94\\
      \makecell[l]{RIFLE \cite{li2020rifle}}          & 89.72$\pm$0.11 & 90.05$\pm$0.26 & 98.70$\pm$0.06 & 87.60$\pm$0.50 & 88.29\\
      \makecell[l]{SCL \cite{gunel2020scl_tune}}      & 89.37$\pm$0.13 & 89.71$\pm$0.20 & 98.65$\pm$0.10 & 87.44$\pm$0.31 & 88.54\\
      \makecell[l]{Bi-tuning \cite{zhong2020bi}}      & 89.41$\pm$0.28 & 89.90$\pm$0.06 & 98.57$\pm$0.10 & 87.39$\pm$0.01 & 88.58\\
      \makecell[l]{Core-tuning \cite{core}}            & 90.17$\pm$0.03 & \textbf{92.36}$\pm$0.14 & 99.18$\pm$0.15 & 89.48$\pm$0.17 & 90.47\\
      \makecell[l]{SSF* \cite{ssf}}                     & 62.22$\pm$0.21 & 84.89$\pm$0.17 & 92.15$\pm$0.55 & 62.38$\pm$0.55 & 81.95\\
      \makecell[l]{\textbf{DR-Tune (Ours)}}           &\textbf{90.60}$\pm$0.15 & 90.57$\pm$0.09 & \textbf{99.27}$\pm$0.10 & \textbf{89.80}$\pm$0.09 & \textbf{91.35} \\ 
                                                      
      \bottomrule 
    \end{tabular}
  \caption{Comparison of the top-1 accuracies (\%) as well as the standard errors by using various fine-tuning methods based on the \emph{self-supervised pretrained model}, \emph{i.e.} ResNet-50 pretrained by MoCo-v2 on ImageNet. `*'  indicates that the method is re-implemented. The best results are in \textbf{bold}.}
  \label{std}
\end{table*}

\begin{table*}[t]\small  
	\centering 
  \setlength{\tabcolsep}{7pt}
	\begin{tabular}{l c c c c c c c c} 
      \toprule 
      \makecell[l]{Method} & CIFAR100$^\dagger$ & Caltech101$^\dagger$ & DTD$^\dagger$ & Flowers$^\dagger$ & Pets$^\dagger$ & SVHN & Sun397 & Avg.\\
      \midrule
      \makecell[l]{Core-tuning \cite{core}}       & 66.3$\pm$0.55 & 89.7$\pm$0.07 & 70.9$\pm$0.03 & 99.0$\pm$0.05 & 92.3$\pm$0.16 & 76.4$\pm$0.08 & 52.5$\pm$0.85 & 78.16 \\
      \makecell[l]{\textbf{DR-Tune (Ours)}}       & \textbf{81.1}$\pm$0.34 & \textbf{92.8}$\pm$0.19 & 71.4$\pm$0.41 & 99.3$\pm$0.02 & \textbf{92.4}$\pm$0.21 & \textbf{92.0}$\pm$0.10 & \textbf{54.5}$\pm$0.03 & \textbf{83.36}\\                                      
      \bottomrule 
    \end{tabular}
  \caption{Comparison of the top-1 accuracies (\%) as well as the standard errors by using various fine-tuning methods based on the {\emph{supervised pretrained model}, \emph{i.e.} ViT-B pretrained on ImageNet.} `*' indicates that the method is re-implemented. `$\dagger$' refers to the training/test split setting as in \cite{vtab}. The best results are in \textbf{bold}.}
  \label{std_2}
\end{table*}

\begin{table}[t] \small
\setlength{\tabcolsep}{6pt}

\centering 
    \begin{tabular}{c c c c} 	
    \toprule
    \makecell[l]{Method} & MPA $\uparrow$ & FWIoU $\uparrow$ & MIoU $\uparrow$ \\
    \midrule
    \makecell[l]{CE-tuning} & 87.31 & 90.26 & 78.42\\
    \makecell[l]{Core-tuning \cite{core}} & 88.76 & 90.75 & 79.62\\
    \makecell[l]{DR-Tune (\textbf{Ours})} & \textbf{89.90} & \textbf{90.81} & \textbf{79.93}\\
    \bottomrule 
    \end{tabular}
\caption{Results (\%) on PASCAL VOC for semantic segmentation, using DeepLab-V3 \cite{deeplab} with ResNet-50 pretrained by MoCo-v2.}
  \label{segment} 
\end{table}

\section{Results on Semantic Segmentation}
\label{segmentation}

In this section, we evaluate the generalizability of DR-Tune on the semantic segmentation task beyond classification.

Following the same setting as \cite{core} does, we evaluate DR-Tune on semantic segmentation. Since only CE-tuning and Core-tuning report the results on this task among the counterparts in Table~1, we take them for comparison. As Table~\ref{segment} displays, DR-Tune clearly outperforms them, showing its generalizability beyond classification.

\begin{table}[t] \small
  \setlength{\tabcolsep}{6pt}
	\centering 

     \begin{tabular}{c c c} 	
	\toprule
    \makecell[l]{Method} & w/o. SC & w. SC (\textbf{Ours}) \\
    \midrule
    \makecell[l]{$\mathrm{MMD}(\mathcal{Z}^{p},\mathcal{Z}^{d})$} & 1.478 & 0.028\\
    \bottomrule 
	\end{tabular}
  \caption{Comparison in terms of MMD on CIFAR10.}
  \vspace{-5pt}
  \label{mmd} 
\end{table}

\section{Qualitative Results}
\label{qual}

\textbf{Visualization of the SC process.}  We provide visualization results on CIFAR10 to demonstrate the effectiveness of the transformations used in the SC module. As displayed in Fig. \ref{visual_sc}, the pretrained feature distribution (low brightness) and the downstream counterpart (high brightness) clearly exhibit a semantic drift. Global rotation mitigates the misalignment of the overall shape as well as the overall center. Class-level translations align the centers for each class, further alleviating the semantic drift. We also add quantitative evaluations by adopting the Maximum Mean Discrepancy (MMD) \cite{mmd} metric in Table \ref{mmd}, showing that the distribution distance remarkably decreases.

\begin{figure}[!t]
  \centering 
  \includegraphics[width=63mm]{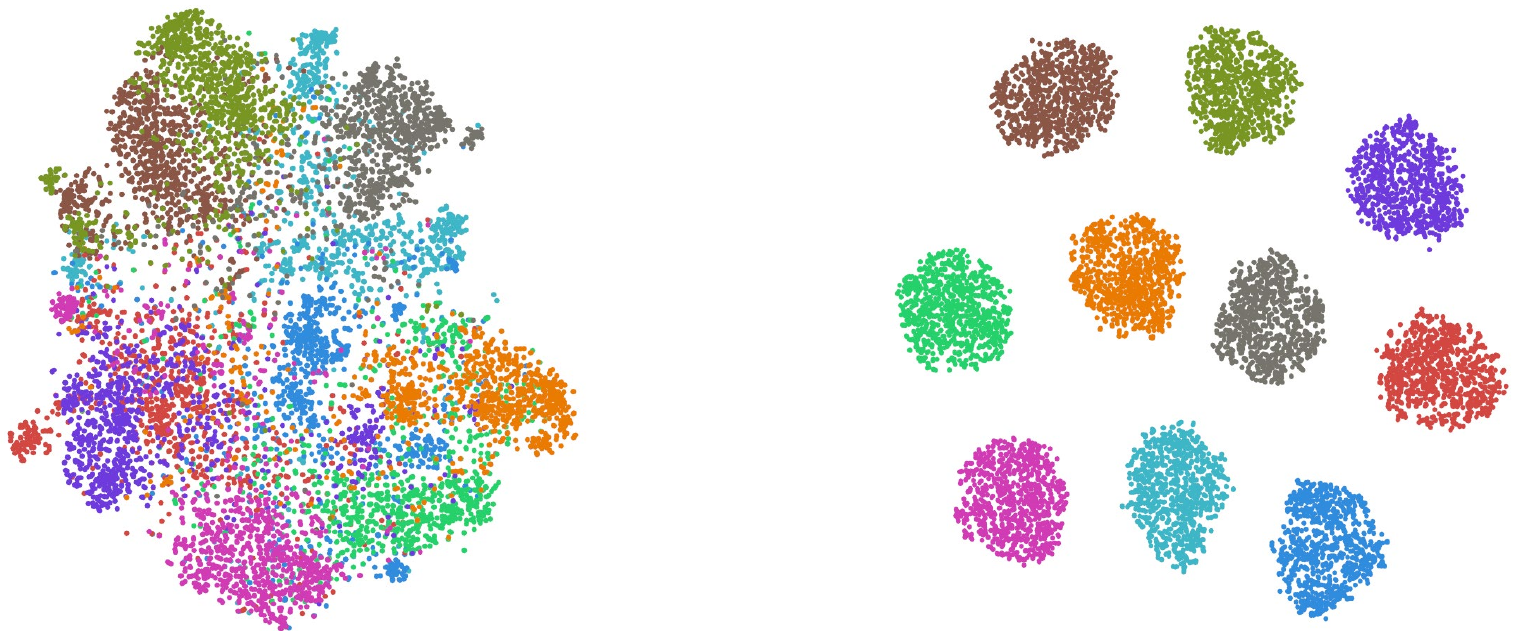}
  \caption{$t$-SNE visualization of distributions of the pretrained (left) and downstream (right) features on CIFAR10.}
  \label{visual_pd}
\end{figure}

\textbf{Visualization of the feature distribution.} In Sec 3.4 of the main body, due to the lack of supervision in the downstream task, the inter-class distribution of the pretrained feature is less discriminative than the downstream one. To make it more convincing, we visualize the distributions of the pretrained and downstream features on CIFAR10 in Fig.~\ref{visual_pd}, where the downstream ones are more discriminative.

\begin{figure*}[!t]
  \centering 
  \subfigure[Results of training samples on CIFAR10.]{
        \includegraphics[width=170mm]{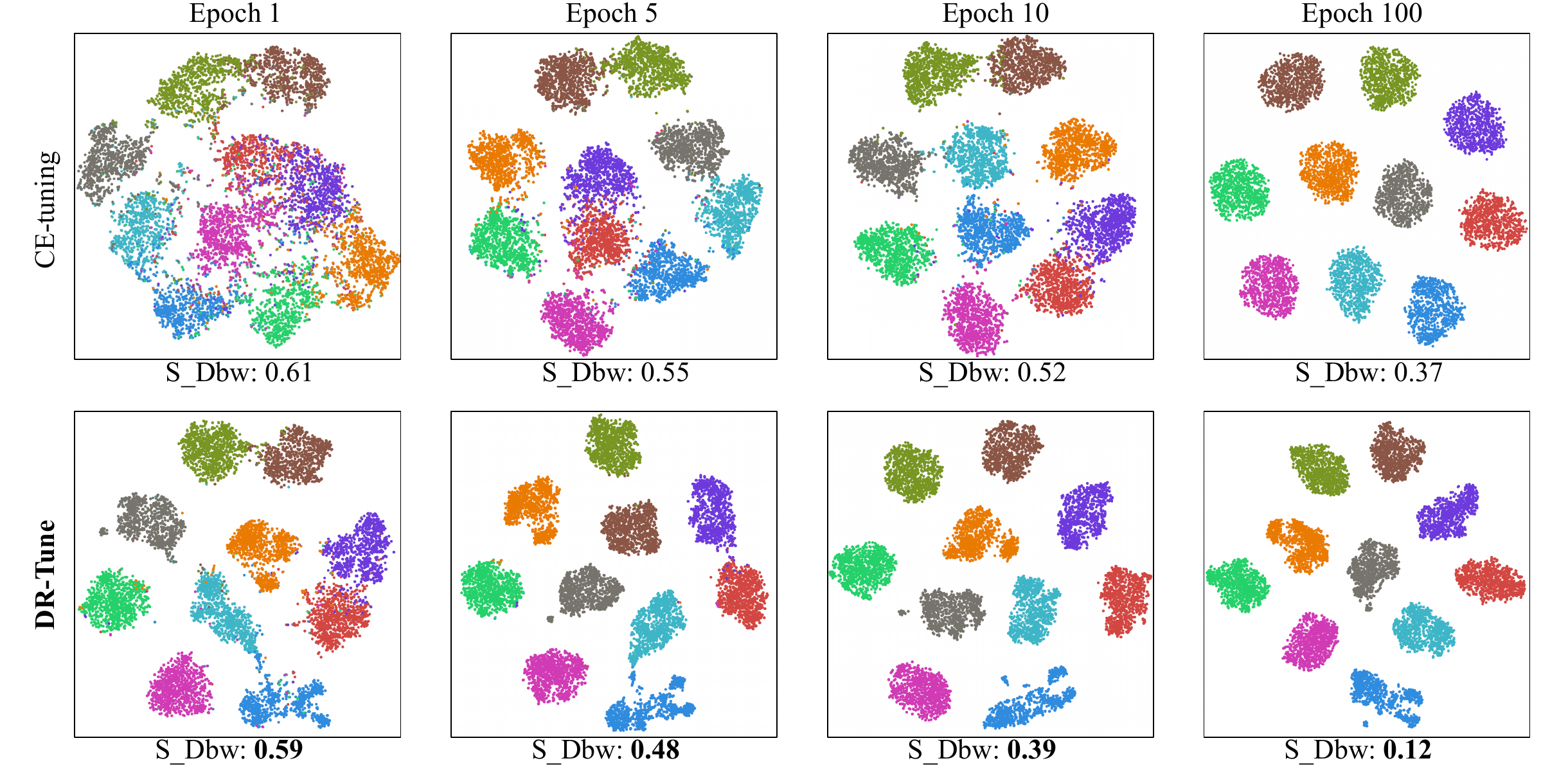}}\\
    \subfigure[Results of testing samples on CIFAR10.]{
            \includegraphics[width=170mm]{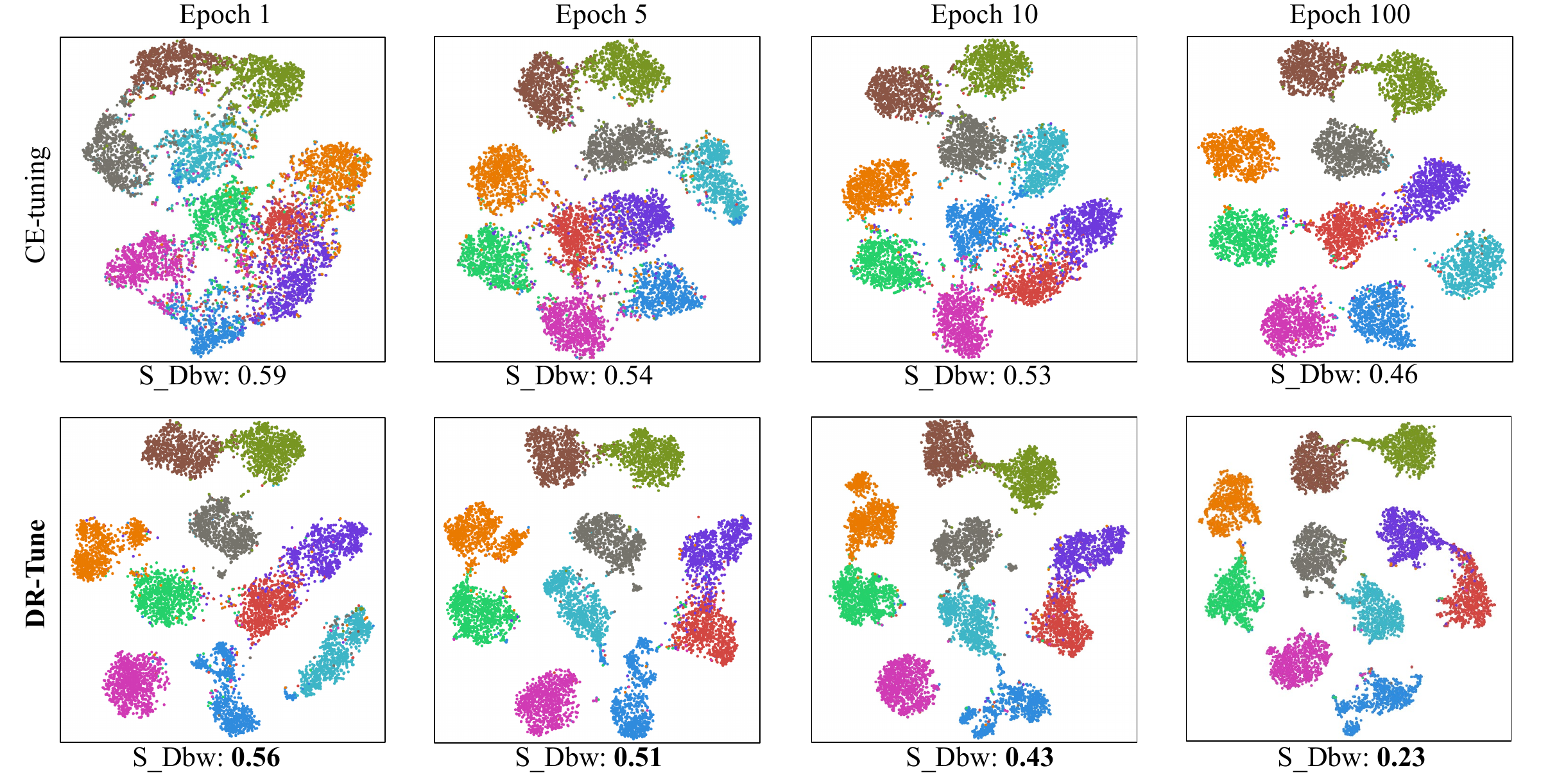}
        }
  \caption{$t$-SNE visualizattion and S\_Dbw scores of the learned features on the CIFAR10 dataset: (a) on the training samples and (b) on the testing samples. CE-tuning refers to vanilla fine-tuning.}
  \label{fast_convergence_test}
\end{figure*}

\textbf{Visualization of the training process.}
In Fig.~\ref{fast_convergence_test}, we use $t$-SNE \cite{tsne} to visualize the features of the training and testing sets from CIFAR10 \cite{cifar} during training. We also use the S\_Dbw score \cite{sdbw} to evaluate the inter-class density and intra-class variance of the learned features where a lower S\_Dbw score is better. DR-Tune utilizes the prior knowledge that accelerates the convergence, and therefore a faster convergence process is observed compared to vanilla fine-tuning (\ie CE-tuning), which only uses the pre-trained model for initialization. Besides, after training, the features obtained by DR-Tune have a lower S\_Dbw score, indicating a more compact intra-class distribution and a more dispersed inter-class distribution.

\section{Limitations}
\label{limit}

As discussed in Sec.~\ref{Analysis}, DR-Tune suffers from a high training latency, due to computation of rotations by SVD in SC, which can be further improved by more efficient solutions. Besides, SC aligns the downstream and pretrained features by a global feature after average pooling for classification, ignoring spatial misalignment, which is crucial to spatio-sensitive tasks, \emph{e.g.} object detection ans semantic segmentation, leaving room for gains.

\end{document}